\pdfoutput=1

\documentclass[11pt]{article}

\usepackage[preprint]{acl}

\usepackage{times}
\usepackage{algorithm}
\usepackage{algpseudocode}
\usepackage{arydshln}
\usepackage{amsmath}
\usepackage{amssymb}
\usepackage{amsfonts}
\usepackage{bm}
\usepackage{bbm}
\usepackage{url}
\usepackage{booktabs}
\usepackage{multirow}
\usepackage{graphicx}
\usepackage{tabularx}
\usepackage{xcolor} 
\usepackage{siunitx}
\usepackage{booktabs}
\usepackage{etoolbox}
\usepackage{caption}
\usepackage{subcaption}
\usepackage{comment}

\usepackage[subrefformat=parens]{subcaption}
	
\usepackage{color, colortbl}

\definecolor{LightCyan}{rgb}{0.88,1,1}

\usepackage[T1]{fontenc}

\usepackage[utf8]{inputenc}

\usepackage{microtype}

\usepackage{inconsolata}

\usepackage{graphicx}

\title{Are Data Augmentation Methods in Named Entity Recognition Applicable for Uncertainty Estimation?}

\author{Wataru Hashimoto, \
  Hidetaka Kamigaito, \
  Taro Watanabe \\
  Nara Institute of Science and Technology \\
  \texttt{\{hashimoto.wataru.hq3, kamigaito.h, taro\}@is.naist.jp}}

\begin{document}
\maketitle
\begin{abstract}
This work investigates the impact of data augmentation on confidence calibration and uncertainty estimation in Named Entity Recognition (NER) tasks. For the future advance of NER in safety-critical fields like healthcare and finance, it is essential to achieve accurate predictions with calibrated confidence when applying Deep Neural Networks (DNNs), including Pre-trained Language Models (PLMs), as a real-world application. However, DNNs are prone to miscalibration, which limits their applicability. Moreover, existing methods for calibration and uncertainty estimation are computational expensive. Our investigation in NER found that data augmentation improves calibration and uncertainty in cross-genre and cross-lingual setting, especially in-domain setting. Furthermore, we showed that the calibration for NER tends to be more effective when the perplexity of the sentences generated by data augmentation is lower, and that increasing the size of the augmentation further improves calibration and uncertainty.
\end{abstract}

\section{Introduction}

Named Entity Recognition (NER) is a one of the fundamental tasks in Natural Language Processing (NLP) to find mentions of named entities and classify them into predefined categories. The predicted information by NER is essential for downstream tasks like event detection \cite{VAVLIAKIS20131}, information retrieval \cite{ner-search}, and masking of personal user information \cite{mask-pui}. Due to the demand, NER is the underlying technology for information extraction from text and documents.

Based on the recent advances in Deep Neural Networks (DNNs), NER's performance is also improved like other NLP fields.
In recent years, Pre-trained Language Models (PLMs) based architectures, such as BERT \cite{devlin-etal-2019-bert} and DeBERTa \cite{he2021deberta}, have been strong baselines in many NLP tasks, including NER.

%
In general, however, DNNs are prone to miscalibration \cite{pmlr-v70-guo17a}, including PLMs \cite{desai-durrett-2020-calibration}; \textit{calibration} means the predicted confidence of the model aligns with the accuracy.\footnote{For example, a predicted confidence of 0.70 from a \textit{perfectly calibrated} network should be 70$\%$ accuracy for that inputs.} The problem causes DNNs to make incorrect predictions with high confidence, which limits the applicability of DNNs on the number of domains where the cost of errors is high, e.g., healthcare and finance. Therefore, DNNs need to provide high prediction performance with appropriately calibrated confidence at the same time.

Confidence calibration and uncertainty estimation methods are ways to solve the miscalibration of DNNs, and have been applied in NLP tasks such as text classification \cite{xiao-2019-uncertainty}, structured prediction \cite{jiang-etal-2022-calibrating, reich-etal-2020-ensemble}, question answering \cite{si-etal-2022-examining}, and machine translation \cite{malinin2021uncertainty}. However, many methods for confidence calibration and uncertainty estimation, typically Monte-Carlo Dropout (MC Dropout) \cite{pmlr-v48-gal16}, are computationally expensive due to multiple stochastic inferences, making them difficult for real-world application. \par

Data augmentation has also been applied for NER \cite{dai-adel-2020-analysis, zhou-etal-2022-melm}, though, it was focusing on the generalization ability on low-resource data. In computer vision (CV) areas, data augmentation makes the model more robust to the input and leads to confidence calibrations \cite{wen2021combining, Liu_2023_CVPR}, in which the same labels are trained on different representations of the input than the original data.
Based on the findings of these previous studies, there is a possibility that data augmentation in NER can improve confidence calibration without increasing inference time, in contrast to the conventional confidence calibration and uncertainty estimation methods.


In this study, we conducted comprehensive experiments to analyze the impact of data augmentation methods for NER~\cite{dai-adel-2020-analysis, zhou-etal-2022-melm} on the confidence calibration and uncertainty in the cross-genre and cross-lingual settings on OntoNotes 5.0 \cite{pradhan-etal-2013-towards} and MultiCoNER \cite{malmasi-etal-2022-multiconer}, respectively. 


Our experiments yield several findings. First, some data augmentation methods in NER lead to improved confidence calibration and uncertainty estimation, especially in-domain. In particular, entity-prediction-based data augmentation~\cite{zhou-etal-2022-melm} and entity replacement from the same entity type~\cite{dai-adel-2020-analysis} show good performance. On the other hand, common confidence calibration methods, MC Dropout or TS~\cite{pmlr-v70-guo17a} have worse confidence calibration and uncertainty estimation performance than the data augmentation methods in NER, even though the data augmentation methods do not aim to improve confidence calibration and uncertainty estimation. Moreover,  increasing the augmentation size improves performance in confidence calibration and uncertainty estimation. The improvement tends to be better the lower the perplexity of the sentences generated by the data augmentation. Our code is available on \url{https://github.com/wataruhashimoto52/ner_da_uncertainty}.

\section{Related Work}

\paragraph{Named Entity Recognition} In the last decade, NER using DNNs has been widely successful; \citet{lample-etal-2016-neural} reported a sequence-labeling model combining bi-directional LSTM with CRF (BiLSTM-CRF). \citet{akbik-etal-2018-contextual} proposed contextualized character-level word embeddings combined with BiLSTM-CRF. In recent years, NER models based on PLMs, such as BERT \cite{devlin-etal-2019-bert}, RoBERTa \cite{roberta-2019}, and DeBERTa \cite{he2021deberta}, have achieved state-of-the-art performance.

\paragraph{Uncertainty Estimation}
In general, DNNs are prone to miscalibration and overconfidence~\cite{pmlr-v70-guo17a} especially without pretraining~\cite{desai-durrett-2020-calibration, ulmer-etal-2022-exploring}.
One way to estimate uncertainty is to run multiple stochastic predictions. Deep Ensemble \cite{deep-ensemble-2017} trains multiple DNN models and integrates their multiple stochastic predictions to make a final prediction. MC Dropout \cite{pmlr-v48-gal16} applies Dropout \cite{JMLR:v15:srivastava14a} regularization at both training and inference time, and by taking multiple samples of the network outputs during inference. These are known to perform calibration well in many cases \cite{NEURIPS2019_eval_uncertainty, pmlr-v130-immer21a}, but their practical use is hampered by the fact that they make multiple probabilistic predictions. A relatively lightweight calibration method is the post-hoc approach. For example, temperature scaling \cite{pmlr-v70-guo17a} performs calibration via dividing logits by a constant, which is a simple and lightweight baseline.

\paragraph{Data Augmentation} Data augmentation methods are widely used in machine learning, CV, and NLP areas. More recent attention has focused on the provision of data augmentation methods to improve calibration and uncertainty. Test-time augmentation (TTA) \cite{Ashukha2020Pitfalls} generates multiple samples during inference and integrates the predictions to estimate the prediction uncertainty. MixUp \cite{zhang2018mixup} uses linear interpolation between two samples to augment a new sample with soft labels, which has been investigated for situations where it is effective for calibration \cite{pmlr-v162-zhang22f}.

In NLP tasks, the impact of data augmentation on calibration in text classification has been investigated in recent study \cite{kim-etal-2023-bag}, but only for In-domain (ID) and not for NER. Furthermore, it has been found that predictive performance is driven by data augmentation in NER \cite{dai-adel-2020-analysis, chen-etal-2020-local, zhou-etal-2022-melm, chen-etal-2022-style, hu-etal-2023-entity}, but these studies have focused only on the predictive performance of NER and have not evaluated for calibration and uncertainty. This is the first study to comprehensively investigate the impact of data augmentation on calibration and uncertainty in NER, both in ID and OOD (Out-of-domain) settings.

\section{Methods}
\label{sec:methods}
In this section, we describe the popular baseline methods for confidence calibration and data augmentation methods for NER. Details about existing calibration methods are described in Appendix \ref{sec:appendix_details_existing_baseline_methods}.

\subsection{Existing Calibration Methods}
\label{sec:baseline_methods}
\paragraph{Baseline} Baseline uses the maximum probability from the softmax layer.

\paragraph{Temperature Scaling (TS)}
TS \cite{pmlr-v70-guo17a} is a post-processing technique for calibrating the confidence scores outputted by a neural network. It involves scaling the logits (i.e., the outputs of the final layer before the softmax) by a temperature parameter $T$ before applying the softmax function to obtain the calibrated probabilities. 

\paragraph{Label Smoothing (LS)}
LS \cite{Miller1996AGO, pereyra2017regularizing} is prevalent regularization technique in machine learning, introduces a controlled level of uncertainty into the training process by modifying the cross-entropy loss.

\paragraph{Monte-Carlo Dropout (MC Dropout)}
MC Dropout is a regularization technique that can be used for uncertainty estimation in neural networks, which requires multiple stochastic inferences \cite{pmlr-v48-gal16}. We perform 20 stochastic inferences and output their average.

\subsection{Data Augmentation Methods for NER}
We investigate data augmentation methods in NER \cite{dai-adel-2020-analysis, zhou-etal-2022-melm} for confidence calibration and uncertainty estimation.
\label{sec:data_augmentation_methods}
\paragraph{Label-wise Token Replacement (LwTR)} LwTR uses binomial distribution to determine whether a token is replaced.
The chosen token is randomly replaced with another token with the same label based on label-wise token distribution on training data. Thus, LwTR keeps the original label sequence.

\paragraph{Mention Replacement (MR)} Unlike LwTR, MR replaces an entity with another entity with the same label instead of a token. Other parts are the same as LwTR. Since entities can have multiple tokens, MR does not keep the original label sequence.

\paragraph{Synonym Replacement (SR)} SR is similar to LwTR except that SR replaces a token with its synonym in WordNet \cite{miller1995wordnet}. Since the synonym can have multiple tokens, SR does not keep the original label sequence.

\paragraph{Masked Entity Language Modeling (MELM)} MELM \cite{zhou-etal-2022-melm} performs data augmentation using a language model that predicts contextually appropriate entities for sentences in which entity parts are masked by entity markers.

\section{Evaluation Metrics}
\label{sec:evaluation_metrics}
We use Expected Calibration Error (ECE), Maximum Calibration Error (MCE), and Area Under Precision-Recall Curve (AUPRC) to evaluate confidence calibration and uncertainty estimation.

\subsection{Expected Calibration Error (ECE)}
ECE \cite{aaai-2015-ece} measures the difference between the accuracy and confidence of a model. Specifically, it calculates the difference between the average confidence and the actual accuracy of the model on different confidence levels. Formally, ECE is defined as:
\begin{equation*}
    \text{ECE} = \sum_{b=1}^B \frac{|\mathcal{D}_b|}{n} \left \lvert \text{acc}( \mathcal{D}_b) - \text{conf}( \mathcal{D}_b) \right \rvert 
\end{equation*}
where $B$ is the number of confidence interval bins, $\mathcal{D}_b$ is the set of examples whose predicted confidence scores fall in the $b$-th interval, $n$ is the total number of examples, $\text{acc}(\mathcal{D}_b)$ is the accuracy of the model on the examples in $\mathcal{D}_b$, and $\text{conf}(\mathcal{D}_b)$ is the average confidence of the model on the examples in $\mathcal{D}_b$.

\subsection{Maximum Calibration Error (MCE)}
MCE \cite{aaai-2015-ece} is the maximum difference between the accuracy and the confidence of the model on different confidence levels. Formally, MCE is defined as:
\begin{equation*}
    \text{MCE} = \max_{b=1}^B \left\lvert \text{acc}(\mathcal{D}_b) - \text{conf}(\mathcal{D}_b) \right\rvert,
\end{equation*}
MCE takes the maximum calibration error in each bin, not the expectation; a smaller MCE means that the model's predictions are less likely to be far off in a given confidence region.

\subsection{Area Under the Precision-Recall Curve (AUPRC)}
AUPRC is the summary statistic the relationship between precision and recall at different thresholds. The higher the value, the higher the overall precision at a given threshold.

\section{Experimental Settings}
\subsection{Datasets}
We conducted experiments on two different NER datasets to evaluate the performance of confidence calibration methods in different settings. For the cross-genre evaluation, we used the OntoNotes 5.0 dataset \cite{pradhan-etal-2013-towards}, which consists of six different genres, broadcast conversation ($\mathtt{bc}$), broadcast news ($\mathtt{bn}$), magazine ($\mathtt{mz}$), newswire ($\mathtt{nw}$), telephone conversation ($\mathtt{tc}$), and web data ($\mathtt{wb}$). This dataset is commonly used for NER evaluation in a cross-domain setting \cite{chen-etal-2021-data}.

For the cross-lingual evaluation, we used the MultiCoNER dataset, which is a large multilingual NER dataset from Wikipedia sentences, questions, and search queries~\cite{malmasi-etal-2022-multiconer}. We selected English as the source language and English, German, Spanish, Hindi, and Bangla as the target languages. The details of the dataset statistics are provided in Table \ref{tab:datasets}.

\begin{table}
\centering
\scalebox{0.85}{
\begin{tabular}{llcccc}
\hline
Dataset $\&$ Domain & $N_{ent}$ & Train & Dev & Test \\
\hline 
$\textbf{OntoNotes 5.0}$ \\
 $\mathtt{bc}$ & 18 & 11,866 & 2,117 & 2,211 \\
 $\mathtt{bn}$ & 18 & 10,683 & 1,295 & 1,357 \\
 $\mathtt{mz}$ & 18 & 6,911 & 642 & 780 \\
 $\mathtt{nw}$ & 18 & 33,908 & 5,771 & 2,197 \\
 $\mathtt{tc}$ & 18 & 11,162 & 1,634 & 1,366 \\
 $\mathtt{wb}$ & 18 & 7,592 & 1,634 & 1,366 \\\hline
$\textbf{MultiCoNER}$ \\
 English ($\mathtt{EN}$) & 6 & 15,300 & 800 & 10,000 \\
 German ($\mathtt{DE}$) & 6 & - & - & 10,000 \\
 Spanish ($\mathtt{ES}$) & 6 & - & - & 10,000 \\
 Hindi ($\mathtt{HI}$) & 6 & - & - & 10,000 \\ \hline
\end{tabular}
}
\caption{Dataset statistics. The table presents the number of entity types, and sequences for the train, development, and test parts of the datasets. For MultiCoNER, we randomly sampled and fixed 10,000 cases out of 200,000 test cases.}
\label{tab:datasets}
\end{table}

\begin{table}
\centering
\scalebox{0.82}{
\begin{tabular}{llcccc}
\hline
Dataset $\&$ Domain & LwTR & MR & SR & MELM ($\eta$, $\mu$) \\
\hline 
$\textbf{OntoNotes 5.0}$ \\
 $\mathtt{bc}$ & 0.3 & 0.7 & 0.3 & (0.5, 0.5) \\
 $\mathtt{bn}$ & 0.4 & 0.8 & 0.2 & (0.7, 0.3) \\
 $\mathtt{mz}$ & 0.7 & 0.4 & 0.5 & (0.3, 0.3) \\
 $\mathtt{nw}$ & 0.7 & 0.5 & 0.7 & (0.7, 0.7) \\
 $\mathtt{tc}$ & 0.4 & 0.4 & 0.1 & (0.3, 0.3) \\
 $\mathtt{wb}$ & 0.7 & 0.7 & 0.8 & (0.5, 0.7) \\\hline
$\textbf{MultiCoNER}$ \\
 English ($\mathtt{EN}$) & 0.2 & 0.8 & 0.4 & (0.3, 0.3) \\ \hline
\end{tabular}
}
\caption{Optimized hyperparameters in data augmentation methods in each source domain. We present the binomial distribution parameters for LwTR, SR and MR, and ($\eta$, $\mu$) for MELM, respectively.}
\label{tab:data_augmentation_hyperparameters}
\end{table}

\subsection{Training Details}
\label{sec:train-details}
In all experiments, we train out models on a single NVIDIA A100 GPU with 40GB of memory. We used MIT-licensed mDeBERTaV3 ($\texttt{microsoft/mdeberta-v3-base}$) \cite{he2023debertav} whose model size is 278M, as a multilingual transformer encoder from Hugging Face $\mathtt{transformers}$~\cite{wolf-etal-2020-transformers} pre-trained model checkpoints, and extracted entities via \textit{sequence labeling}. Cross-entropy loss is minimized by AdamW \cite{adamw-2019} with a linear scheduler \cite{linear-scheduler-2017}. The batch size is 32, and gradient clipping is applied with maximum norm of 1. The initial learning rate was set to 1e-5. To avoid overfitting, we also applied early stopping with $patients=5$.

For the temperature parameter in TS, we used Optuna \cite{optuna-2019} to optimize the temperature parameter based on dev set loss with a search range of [0.001, 0.002, ..., 5.000] in 100 trials. In addition, we optimized the binomial distribution parameter to manipulate replacement intensity for data augmentation methods using the dev set by a grid search in the range of [0.1, 0.2, ..., 0.8]. In LS, we conducted a grid search in the range of [0.01, 0.05, 0.1, 0.2, 0.3] to optimize the smoothing parameter. In the case of MELM, mask rate $\eta$ during fine tuning and mask parameter $\mu$ during generation are hyperparameters. We conducted a grid search for each hyperparameter in the range [0.3, 0.5, 0.7], as in \citet{zhou-etal-2022-melm}. All hyperparameters in data augmentation are shown in Table ~\ref{tab:data_augmentation_hyperparameters}. The implementations of LwTR, MR and SR are used several repos,\footnote{\url{https://github.com/boschresearch/data-augmentation-coling2020}}~\footnote{\url{https://github.com/kajyuuen/daaja}} while the implementation of MELM used the official repo.\footnote{\url{https://github.com/RandyZhouRan/MELM}}

We perform each experiment 10 times using different random seeds, collect evaluation metric values, and report their average and standard deviation. For convenience, the reported values are multiplied by 100.

\begin{table*}[t!]
\centering
\scalebox{0.52}{
\begin{tabular}{l|cc|cc|cc|cc|cc|cc}
\hline
\multicolumn{1}{l|}{Methods} & \multicolumn{2}{c|}{$\mathtt{bc}$} & \multicolumn{2}{c|}{$\mathtt{bn}$} & \multicolumn{2}{c|}{$\mathtt{mz}$} & \multicolumn{2}{c|}{$\mathtt{nw}$} & \multicolumn{2}{c|}{$\mathtt{tc}$} & \multicolumn{2}{c}{$\mathtt{wb}$} \\ \cline{2-13} 
\multicolumn{1}{c|}{} & ECE (↓) & MCE (↓) & ECE (↓) & MCE (↓) & ECE (↓) & MCE (↓) & ECE (↓) & MCE (↓) & ECE (↓) & MCE (↓) & ECE (↓) & MCE (↓) \\ \hline
Baseline   & 18.87$\pm$0.73 & 23.58$\pm$1.01 & 11.50$\pm$0.75  & 16.14$\pm$1.97 & 15.75$\pm$0.94 & 20.93$\pm$0.97 & 11.74$\pm$0.27 & 16.15$\pm$0.77 & 31.17$\pm$1.56 & 33.81$\pm$1.67 & 28.86$\pm$1.51 & 34.38$\pm$1.82\\ 
TS         & 18.86$\pm$0.68 & 23.22$\pm$0.86 & 11.25$\pm$0.55 & 15.43$\pm$1.41 & 15.40$\pm$0.74 & 20.30$\pm$1.23 & 11.71$\pm$0.36 & 15.80$\pm$0.85 & 27.95$\pm$2.51 & 30.70$\pm$2.55 & 29.70$\pm$1.54 & 34.88$\pm$1.66 \\ 
LS         & 19.29$\pm$1.04 & 24.11$\pm$1.57 & 11.42$\pm$0.52 & 15.31$\pm$1.24 & 15.59$\pm$0.85 & 20.91$\pm$1.30 & 12.05$\pm$0.20  & 16.83$\pm$0.36 & 26.46$\pm$1.36 & 28.89$\pm$1.42 & 29.34$\pm$2.25 & 34.86$\pm$2.22 \\ 
MCDropout  & 18.69$\pm$0.71 & 23.54$\pm$1.31 & 11.38$\pm$0.71 & 15.73$\pm$1.60 & 15.89$\pm$0.29 & 21.15$\pm$0.54 & 11.83$\pm$0.55 & 16.56$\pm$1.41 & 29.01$\pm$2.50  & 31.94$\pm$2.81 & 28.41$\pm$1.45 & 33.88$\pm$1.77 \\ \hline
LwTR (DA)  & 19.15$\pm$0.55 & 23.70$\pm$0.77  & 11.72$\pm$0.42 & 16.37$\pm$1.21 & 15.12$\pm$0.44 & 20.56$\pm$0.80 & 11.82$\pm$0.39 & 15.57$\pm$0.47 & 28.78$\pm$2.27 & 31.31$\pm$2.14 & 28.72$\pm$1.70 & 34.30$\pm$1.68 \\ 
MR (DA)    & 19.13$\pm$0.95 & 23.17$\pm$1.10  & 11.59$\pm$0.34 & 15.89$\pm$0.92 & 14.66$\pm$1.05 & 19.63$\pm$1.37 & 11.50$\pm$0.33  & 15.62$\pm$0.74 & 28.65$\pm$3.20  & 31.23$\pm$3.18 & 27.08$\pm$1.40 & 32.39$\pm$1.57 \\ 
SR (DA)    & \textbf{18.16$\pm$0.63} & \textbf{21.99$\pm$0.91\textsuperscript{†}} & 11.38$\pm$0.44 & 15.44$\pm$0.96 & 15.29$\pm$0.96 & 20.11$\pm$1.14 &11.71$\pm$0.25 & 16.31$\pm$0.57 & 27.30$\pm$4.37  & 29.85$\pm$4.54 & 29.72$\pm$0.91 & 34.74$\pm$1.05 \\ 
MELM (DA)  & 18.59$\pm$0.60  & 22.67$\pm$0.95 & \textbf{10.75$\pm$0.46\textsuperscript{†}} & \textbf{14.11$\pm$0.69\textsuperscript{†}} & \textbf{13.94$\pm$0.98\textsuperscript{†}} & \textbf{18.50$\pm$1.22\textsuperscript{†}} & \textbf{11.28$\pm$0.33\textsuperscript{†}} & \textbf{15.43$\pm$0.98} & \textbf{25.71$\pm$1.73} & \textbf{28.19$\pm$1.87} & \textbf{26.58$\pm$1.48\textsuperscript{†}} & \textbf{31.47$\pm$1.64\textsuperscript{†}} \\ \hline
\end{tabular}
}
\caption{Results of existing methods and data augmentation methods in OntoNotes 5.0 in ID setting. The best results are shown in bold. {\textdagger} indicates significantly improved than existing methods (p < 0.05) by using t-test.}
\label{tab:ontonotes5_id_scores}
\end{table*}

\subsection{Evaluation Details}
\label{sec:eval-details}
The NER model calibration is evaluated based on the "Event of Interests" concept introduced in the previous study \cite{NIPS2015-kuleshov, jagannatha-yu-2020-calibrating}. Since the full label space $|\mathcal{Y}|$ is large for structured prediction tasks such as NER, we focus instead on the event set $L(x)$, which is the set containing the events of interest $E \in L(x)$ obtained by processing the model output. 

There are two main strategies for constructing $L(x)$: The first strategy is to construct $L(x)$ only from the events obtained by the MAP label sequence prediction of the model; The second strategy is to construct $L(x)$ from all possible label sequences; The first strategy is easy to obtain events, but the coverage of events is low depending on the model's prediction. The second strategy provides a high coverage of events, but is computationally expensive to obtain events. \citet{jagannatha-yu-2020-calibrating} is based on the first strategy, where the entities extracted by the NER model are calibrated on the basis of forecasters (e.g., gradient boosting decision trees \cite{friedman2000greedy}), which are binary classifiers separate from the NER model. Since the training dataset for forecasters consists of entities extracted by the NER model, more entities are needed to improve the uncertainty performance of the forecasters. Therefore, for example, the top-k Viterbi decoding of the CRF is used to increase the entity coverage and the size of the forecaster's training dataset. \par
On the other hand, \citet{jiang-etal-2022-calibrating} is based on the second strategy, where it introduces a method to find the probability that a span has a specific entity type for datasets with short sequences, such as WikiAnn \cite{pan-etal-2017-cross}, with restricted token sequences and span lengths. However, this method is computationally difficult for datasets with longer token sequences and more complex label spaces, such as OntoNotes 5.0 and MultiCoNER, because the number of spans explodes. We therefore simplify the evaluation process by measuring the calibration of the entity span obtained from the NER model's MAP label sequence prediction of the model. Uncertainty performance is evaluated by taking the product of the probabilities of each token corresponding to an entity as the probability of one entity.

\section{Results and Discussion}
We present the performance of cross-genre and cross-lingual confidence calibration and uncertainty estimation as the main results. The cross-genre evaluations are quantified by learning on a training set in one genre and evaluating calibration and uncertainty on a test in another genre. Similarly, in the cross-lingual evaluations, we train the model in one language (in this research, we use English; $\mathtt{EN}$) and evaluate the calibration and uncertainty on a test set in another language.

\subsection{Cross-genre Evaluation}
\label{sec:cross-genre-calibration}

\begin{table*}[t]
\centering
\scalebox{0.62}{
\begin{tabular}{l|cc|cc|cc|cc|cc}
\hline
\multicolumn{11}{c}{OntoNotes 5.0 ($\mathtt{bc}$)} \\ \hline 
\multicolumn{1}{l|}{Methods} & \multicolumn{2}{c|}{$\mathtt{bn}$} & \multicolumn{2}{c|}{$\mathtt{mz}$} & \multicolumn{2}{c|}{$\mathtt{nw}$} & \multicolumn{2}{c|}{$\mathtt{tc}$} & \multicolumn{2}{c}{$\mathtt{wb}$} \\ \cline{2-11} 
\multicolumn{1}{c|}{} & ECE (↓) & MCE (↓) & ECE (↓) & MCE (↓) & ECE (↓) & MCE (↓) & ECE (↓) & MCE (↓) & ECE (↓) & MCE (↓) \\ \hline
Baseline & 17.54$\pm$0.67 & 25.90$\pm$1.29 & 18.83$\pm$0.89 & 25.65$\pm$1.09 & 23.52$\pm$0.77 & 34.25$\pm$1.41 & \textbf{26.20$\pm$1.23} & \textbf{28.76$\pm$1.30} & 57.47$\pm$0.87 & 62.96$\pm$0.56 \\
TS  & 17.19$\pm$0.81 & 24.93$\pm$1.27 & 19.42$\pm$1.48 & 26.32$\pm$1.97 & 23.51$\pm$1.08 & 33.68$\pm$1.72 & 26.85$\pm$2.11 & 29.36$\pm$2.35 & 57.66$\pm$1.32 & 62.96$\pm$1.15 \\
LS & 17.45$\pm$0.96 & 25.43$\pm$1.77 & 19.38$\pm$1.03 & 26.36$\pm$1.56 & 23.72$\pm$1.01 & 34.23$\pm$1.95 & 26.34$\pm$1.78 & 28.81$\pm$2.04 & \textbf{56.98$\pm$1.17} & \textbf{62.51$\pm$0.91} \\
MC Dropout & 17.50$\pm$0.66 & 25.77$\pm$1.58 & 19.22$\pm$1.21 & 26.39$\pm$1.16 & 23.67$\pm$0.73 & 34.51$\pm$1.59 & 26.32$\pm$1.10 & 28.66$\pm$1.12 & 57.51$\pm$1.29 & 62.80$\pm$0.90 \\ \hline
LwTR (DA) & 17.58$\pm$0.44 & 25.45$\pm$1.34 & 19.34$\pm$1.34 & 26.11$\pm$1.56 & 23.65$\pm$0.53 & 33.89$\pm$1.13 & 27.50$\pm$1.73 & 29.70$\pm$2.01 & 58.68$\pm$1.51 & 63.83$\pm$1.22 \\
MR (DA) & 17.43$\pm$0.62 & 24.99$\pm$1.36 & \textbf{18.38$\pm$1.62} & \textbf{24.93$\pm$1.73} & \textbf{23.28$\pm$0.54} & 33.35$\pm$1.16 & 26.78$\pm$2.19 & 28.85$\pm$2.21 & 59.01$\pm$0.99 & 64.06$\pm$0.76 \\
SR (DA) & \textbf{17.01$\pm$0.39} & \textbf{24.45$\pm$0.74} & 20.01$\pm$1.56 & 26.94$\pm$1.72 & 23.42$\pm$0.66 & \textbf{33.29$\pm$1.33} & 26.62$\pm$1.59 & 28.81$\pm$1.76 & 58.14$\pm$0.79 & 63.02$\pm$0.59 \\
MELM (DA) & 17.22$\pm$0.65 & 24.55$\pm$1.41 & 19.41$\pm$0.80 & 26.01$\pm$1.06 & 23.66$\pm$0.85 & 33.75$\pm$1.46 & 30.11$\pm$1.39 & 32.59$\pm$1.71 & 58.72$\pm$1.42 & 63.71$\pm$1.18 \\
\hline \hline

\multicolumn{11}{c}{OntoNotes 5.0 ($\mathtt{bn}$)} \\ \hline 
\multicolumn{1}{l|}{Methods} & \multicolumn{2}{c|}{$\mathtt{bc}$} & \multicolumn{2}{c|}{$\mathtt{mz}$} & \multicolumn{2}{c|}{$\mathtt{nw}$} & \multicolumn{2}{c|}{$\mathtt{tc}$} & \multicolumn{2}{c}{$\mathtt{wb}$} \\ \cline{2-11} 
\multicolumn{1}{c|}{} & ECE (↓) & MCE (↓) & ECE (↓) & MCE (↓) & ECE (↓) & MCE (↓) & ECE (↓) & MCE (↓) & ECE (↓) & MCE (↓)  \\ \hline
Baseline   & 19.30$\pm$0.82  & 24.37$\pm$1.47 & 20.55$\pm$1.59  & 26.62$\pm$2.55  & 20.05$\pm$0.98  & 28.44$\pm$2.25  & 25.42$\pm$0.73  & 27.56$\pm$0.64  & $\textbf{59.02$\pm$1.16}$  & 63.61$\pm$0.66  \\
TS         & 19.20$\pm$0.88  & 24.18$\pm$1.75  & 21.21$\pm$1.14  & 27.20$\pm$1.72   & 20.34$\pm$0.73  & 28.80$\pm$2.12   & 25.33$\pm$1.28  & 27.57$\pm$1.27  & 59.11$\pm$1.06  & $\textbf{63.60$\pm$0.60}$   \\
LS         & $\textbf{18.37$\pm$0.60}$   & $\textbf{22.52$\pm$1.41}$ & 21.61$\pm$0.47  & 27.04$\pm$1.04  & 19.98$\pm$0.41  & 27.64$\pm$1.11  & $\textbf{24.66$\pm$0.48}$  & $\textbf{26.69$\pm$0.44}$  & 59.92$\pm$0.75  & 63.87$\pm$0.77  \\
MC Dropout  & 18.76$\pm$0.97  & 23.34$\pm$1.56 & 20.91$\pm$0.96  & 26.62$\pm$1.82  & 20.04$\pm$0.57  & 28.25$\pm$1.62  & 25.21$\pm$1.27  & 27.52$\pm$1.17  & 59.09$\pm$0.99  & 63.63$\pm$0.54  \\ \hline
LwTR (DA)  & 20.30$\pm$0.87   & 25.42$\pm$1.18  & 20.71$\pm$1.01  & 27.14$\pm$1.16  & 20.51$\pm$0.41  & 29.04$\pm$1.26  & 26.36$\pm$2.08  & 28.67$\pm$2.09  & 59.32$\pm$0.97  & 64.00$\pm$0.55   \\
MR (DA)    & 19.78$\pm$1.26  & 24.35$\pm$1.85 & 20.19$\pm$0.47  & 26.08$\pm$1.07  & 20.42$\pm$0.60   & 27.83$\pm$1.74  & 25.69$\pm$0.77  & 27.75$\pm$0.81  & 59.57$\pm$0.96  & 64.13$\pm$0.50   \\
SR (DA)    & 19.61$\pm$0.97  & 24.08$\pm$1.64 & $\textbf{19.79$\pm$0.75}$  & $\textbf{25.52$\pm$1.22}$  & 19.81$\pm$0.39  & 27.18$\pm$1.30   & 26.20$\pm$1.56   & 28.42$\pm$1.68  & 59.86$\pm$0.67  & 63.66$\pm$0.40   \\
MELM (DA)  & 19.93$\pm$0.69  & 23.98$\pm$1.09 & 20.40$\pm$0.65   & 25.54$\pm$1.19  & $\textbf{19.73$\pm$0.65}$  & $\textbf{26.80$\pm$1.19}\textsuperscript{†}$ & 28.47$\pm$2.14  & 30.59$\pm$2.15  & 60.51$\pm$0.57  & 64.44$\pm$0.33  \\ \hline \hline

\multicolumn{11}{c}{OntoNotes 5.0 ($\mathtt{nw}$)} \\ \hline
\multicolumn{1}{l|}{Methods} & \multicolumn{2}{c|}{$\mathtt{bc}$} & \multicolumn{2}{c|}{$\mathtt{bn}$} & \multicolumn{2}{c|}{$\mathtt{mz}$} & \multicolumn{2}{c|}{$\mathtt{tc}$} & \multicolumn{2}{c}{$\mathtt{wb}$} \\ \cline{2-11} 
\multicolumn{1}{c|}{} & ECE (↓) & MCE (↓) & ECE (↓) & MCE (↓) & ECE (↓) & MCE (↓) & ECE (↓) & MCE (↓) & ECE (↓) & MCE (↓) \\ \hline
Baseline & \textbf{20.65$\pm$1.79} & \textbf{25.32$\pm$2.15} & \textbf{15.24$\pm$0.65} & 21.06$\pm$1.21 & 22.67$\pm$1.24 & 28.48$\pm$2.17 & 27.81$\pm$1.26 & 30.21$\pm$1.39 & 60.28$\pm$1.17 & 64.30$\pm$0.86 \\
TS & 21.08$\pm$0.75 & 25.80$\pm$1.01 & 15.61$\pm$0.46 & 21.63$\pm$0.80 & 22.76$\pm$1.01 & 28.92$\pm$1.47 & 28.02$\pm$1.61 & 30.21$\pm$1.90 & 60.37$\pm$0.89 & 64.61$\pm$0.68 \\
LS & 20.46$\pm$1.23 & 24.63$\pm$2.21 & 15.51$\pm$0.55 & \textbf{20.80$\pm$1.70} & 22.66$\pm$1.10 & 28.35$\pm$1.86 & 28.50$\pm$1.52 & 30.41$\pm$1.21 & 60.17$\pm$1.05 & 64.07$\pm$0.72 \\
MC Dropout & 21.25$\pm$1.84 & 25.98$\pm$2.09 & 15.58$\pm$0.98 & 21.59$\pm$1.71 & 22.38$\pm$1.10 & 28.34$\pm$1.67 & 28.05$\pm$1.70 & 30.19$\pm$1.79 & 60.64$\pm$0.94 & 64.63$\pm$0.57 \\ \hline
LwTR (DA) & 21.87$\pm$0.87 & 26.58$\pm$0.99 & 15.81$\pm$0.30 & 21.93$\pm$0.41 & 22.76$\pm$0.93 & 28.38$\pm$0.92 & \textbf{27.60$\pm$0.72} & \textbf{29.48$\pm$0.45} & \textbf{59.96$\pm$0.46} & \textbf{64.06$\pm$0.40} \\
MR (DA) & 21.70$\pm$0.27 & 26.29$\pm$0.30 & 15.55$\pm$0.87 & 21.38$\pm$2.16 & \textbf{21.08$\pm$1.21} & \textbf{26.33$\pm$2.14} & 30.35$\pm$2.69 & 32.44$\pm$2.82 & 61.16$\pm$1.06 & 65.12$\pm$0.80 \\
SR (DA) & 21.29$\pm$1.37 & 25.82$\pm$1.31 & 16.00$\pm$0.58 & 21.72$\pm$0.22 & 21.83$\pm$0.67 & 27.37$\pm$0.85 & 33.41$\pm$5.50 & 35.59$\pm$5.44 & 60.58$\pm$0.72 & 64.50$\pm$0.54 \\
MELM (DA) & 21.96$\pm$1.31 & 26.91$\pm$1.88 & 15.83$\pm$0.84 & 21.76$\pm$1.63 & 21.16$\pm$1.38 & 26.88$\pm$1.49 & 33.92$\pm$4.15 & 36.39$\pm$4.03 & 60.94$\pm$0.62 & 65.03$\pm$0.33 \\ \hline \hline

\multicolumn{11}{c}{OntoNotes 5.0 ($\mathtt{tc}$)} \\ \hline
\multicolumn{1}{l|}{Methods} & \multicolumn{2}{c|}{$\mathtt{bc}$} & \multicolumn{2}{c|}{$\mathtt{bn}$} & \multicolumn{2}{c|}{$\mathtt{mz}$} & \multicolumn{2}{c|}{$\mathtt{nw}$} & \multicolumn{2}{c}{$\mathtt{wb}$} \\ \cline{2-11} 
\multicolumn{1}{c|}{} & ECE (↓) & MCE (↓) & ECE (↓) & MCE (↓) & ECE (↓) & MCE (↓) & ECE (↓) & MCE (↓) & ECE (↓) & MCE (↓) \\ \hline
Baseline & 36.70$\pm$1.65 & 44.25$\pm$1.66 & 35.47$\pm$2.48 & 45.75$\pm$2.46 & 37.15$\pm$1.77 & 47.34$\pm$1.79 & 39.08$\pm$0.56 & 52.50$\pm$1.41 & \textbf{46.38$\pm$1.28} & \textbf{54.29$\pm$1.37} \\
TS & 35.69$\pm$2.21 & 43.34$\pm$2.18 & 34.15$\pm$2.65 & 44.48$\pm$2.56 & 36.38$\pm$1.79 & 46.71$\pm$1.43 & 38.59$\pm$1.53 & 52.58$\pm$1.38 & 47.20$\pm$0.92 & 55.31$\pm$1.10 \\
LS & 33.91$\pm$1.86 & 41.50$\pm$1.75 & 31.40$\pm$2.35 & 41.24$\pm$2.43 & 34.14$\pm$1.91 & 44.37$\pm$1.42 & 37.04$\pm$2.25 & 50.00$\pm$1.92 & 48.48$\pm$1.29 & 56.10$\pm$0.89 \\
MC Dropout & 35.83$\pm$2.02 & 43.93$\pm$1.75 & 33.87$\pm$2.02 & 44.31$\pm$1.92 & 36.18$\pm$2.43 & 46.31$\pm$2.43 & 38.97$\pm$0.83 & 52.80$\pm$1.08 & 46.92$\pm$2.04 & 54.95$\pm$2.13 \\ \hline
LwTR (DA) & 34.94$\pm$2.42 & 43.20$\pm$1.90 & 32.61$\pm$3.16 & 43.28$\pm$2.55 & 34.44$\pm$1.83 & 44.98$\pm$1.88 & 37.85$\pm$2.13 & 52.09$\pm$1.60 & 46.78$\pm$1.26 & 54.94$\pm$1.84 \\
MR (DA) & 35.18$\pm$2.89 & 42.62$\pm$2.30 & 33.50$\pm$3.77 & 42.66$\pm$3.20 & 34.35$\pm$2.78 & 44.78$\pm$2.69 & 37.97$\pm$2.64 & 50.85$\pm$3.46 & 48.61$\pm$1.70 & 55.78$\pm$1.90 \\
SR (DA) & 34.58$\pm$2.40 & 42.51$\pm$1.55 & 32.66$\pm$4.13 & 42.57$\pm$3.28 & \textbf{32.69$\pm$3.21} & 43.01$\pm$2.83 & 38.50$\pm$1.51 & 52.00$\pm$1.56 & 46.99$\pm$1.27 & 54.86$\pm$1.40 \\
MELM (DA) & \textbf{33.05$\pm$1.75} & \textbf{40.55$\pm$2.16} & \textbf{29.46$\pm$1.55\textsuperscript{†}} & \textbf{37.81$\pm$1.56\textsuperscript{†}} & 33.46$\pm$1.66 & \textbf{42.78$\pm$2.55} & \textbf{36.79$\pm$1.27} & \textbf{49.33$\pm$2.26} & 50.52$\pm$1.10 & 57.27$\pm$1.27 \\ \hline
\end{tabular}
}
\caption{Results of existing methods and data augmentation methods in OntoNotes 5.0 in OOD test dataset.}
\label{tab:ontonotes5_ood_scores}
\end{table*}

\begin{table*}[t]
\centering
\scalebox{0.53}{
\begin{tabular}{l|c|c|c|c|c|c||c|c|c|c|c|c}
\hline
\multicolumn{1}{l|}{Methods} & \multicolumn{6}{c||}{OntoNotes 5.0 ($\mathtt{bc}$)} 
 & \multicolumn{6}{c}{OntoNotes 5.0 ($\mathtt{bn}$)} \\ \cline{2-13}
\multicolumn{1}{c|}{} & $\mathtt{bc}$ & $\mathtt{bn}$ & $\mathtt{mz}$ & $\mathtt{nw}$ & $\mathtt{tc}$ & $\mathtt{wb}$ & $\mathtt{bc}$ & $\mathtt{bn}$ & $\mathtt{mz}$ & $\mathtt{nw}$ & $\mathtt{tc}$ & $\mathtt{wb}$ \\ \hline
Baseline  & 94.72$\pm$0.21 & 95.13$\pm$0.43 & 96.40$\pm$0.40 & 93.27$\pm$0.41 & 92.69$\pm$0.57 & 93.03$\pm$0.56 & \textbf{95.12$\pm$0.30} & 97.23$\pm$0.20 & 95.83$\pm$0.45 & 95.29$\pm$0.27 & 93.62$\pm$0.59 & 93.13$\pm$0.40 \\
TS        & \textbf{94.89$\pm$0.59} & 95.14$\pm$0.35 & 96.15$\pm$0.51 & 93.26$\pm$0.45 & 92.78$\pm$1.01 & 92.97$\pm$0.83 & 95.05$\pm$0.39 & \textbf{97.38$\pm$0.17} & 95.33$\pm$0.31 & 95.23$\pm$0.20 & \textbf{93.96$\pm$0.51} & \textbf{93.25$\pm$0.29} \\
LS        & 94.74$\pm$0.54 & 95.09$\pm$0.37 & 96.15$\pm$0.30 & 93.15$\pm$0.43 & 92.60$\pm$0.79 & 92.73$\pm$0.36 & 94.99$\pm$0.22 & 97.32$\pm$0.20 & 95.60$\pm$0.22 & 95.11$\pm$0.37 & 93.49$\pm$0.43 & 92.90$\pm$0.47 \\
MC Dropout & 94.71$\pm$0.31 & 95.09$\pm$0.18 & 96.07$\pm$0.24 & 93.11$\pm$0.43 & 92.76$\pm$0.67 & 92.88$\pm$0.33 & 95.03$\pm$0.34 & 97.30$\pm$0.18 & 95.78$\pm$0.46 & 95.29$\pm$0.19 & 93.80$\pm$0.44 & 93.22$\pm$0.35 \\ \hline
LwTR (DA) & 94.53$\pm$0.28 & 95.02$\pm$0.37 & 96.22$\pm$0.33 & 93.23$\pm$0.23 & 92.76$\pm$0.64 & 92.91$\pm$0.52 & 94.36$\pm$0.54 & 97.29$\pm$0.14 & 95.74$\pm$0.16 & 95.15$\pm$0.20 & 93.64$\pm$0.51 & 93.08$\pm$0.49 \\
MR (DA)   & 94.44$\pm$0.29 & 94.88$\pm$0.24 & \textbf{96.53$\pm$0.43} & \textbf{93.4$\pm$0.29} & \textbf{92.82$\pm$0.60} & 92.74$\pm$0.42 & 94.57$\pm$0.50 & 97.20$\pm$0.19 & \textbf{96.27$\pm$0.31\textsuperscript{†}} & 95.11$\pm$0.22 & 93.64$\pm$0.55 & 92.91$\pm$0.52 \\
SR (DA)   & 94.44$\pm$0.35 & 95.09$\pm$0.32 & 95.70$\pm$0.40 & 93.21$\pm$0.37 & 93.24$\pm$0.43 & \textbf{93.06$\pm$0.39} & 94.76$\pm$0.65 & 97.28$\pm$0.15 & 95.85$\pm$0.33 & 95.30$\pm$0.17 & 93.78$\pm$0.63 & 93.06$\pm$0.24 \\
MELM (DA) & 94.51$\pm$0.16 & \textbf{95.15$\pm$0.34}& 96.01$\pm$0.29 & 93.09$\pm$0.44 & 92.64$\pm$0.52 & 92.90$\pm$0.47 & 94.34$\pm$0.47 & 97.24$\pm$0.21 & 96.18$\pm$0.32 & \textbf{95.32$\pm$0.32} & 93.51$\pm$0.50 & 92.97$\pm$0.48 \\ \hline \hline
\multicolumn{1}{l|}{Methods} & \multicolumn{6}{c||}{OntoNotes 5.0 ($\mathtt{nw}$)} 
 & \multicolumn{6}{c}{OntoNotes 5.0 ($\mathtt{tc}$)} \\ \cline{2-13}
\multicolumn{1}{c|}{} & $\mathtt{bc}$ & $\mathtt{bn}$ & $\mathtt{mz}$ & $\mathtt{nw}$ & $\mathtt{tc}$ & $\mathtt{wb}$ & $\mathtt{bc}$ & $\mathtt{bn}$ & $\mathtt{mz}$ & $\mathtt{nw}$ & $\mathtt{tc}$ & $\mathtt{wb}$ \\ \hline
Baseline  & 94.60$\pm$0.80 & 96.36$\pm$0.32 & 95.22$\pm$0.48 & 97.81$\pm$0.12 & 93.32$\pm$0.44 & 93.29$\pm$0.46 & 87.10$\pm$1.25 & 89.22$\pm$0.71 & 84.94$\pm$1.61 & 81.28$\pm$2.58 & 93.45$\pm$0.77 & 89.62$\pm$1.10 \\
TS        & 94.50$\pm$0.40 & \textbf{96.36$\pm$0.32} & 95.34$\pm$0.39 & 97.74$\pm$0.18 & 93.15$\pm$0.52 & 93.33$\pm$0.37 & \textbf{87.74$\pm$1.12} & 89.45$\pm$0.47 & 85.95$\pm$1.65 & 82.50$\pm$1.35 & 93.11$\pm$0.98 & 89.93$\pm$0.88 \\
LS        & \textbf{94.65$\pm$0.30} & 96.23$\pm$0.24 & 95.19$\pm$0.57 & 97.70$\pm$0.09 & 93.05$\pm$0.43 & 93.39$\pm$0.41 & 87.07$\pm$1.00 & 89.57$\pm$0.76 & \textbf{86.67$\pm$1.75} & 82.79$\pm$1.09 & 92.75$\pm$1.06 & 90.66$\pm$0.61 \\
MC Dropout & 94.37$\pm$0.92 & 96.32$\pm$0.23 & 95.27$\pm$0.31 & 97.81$\pm$0.24 & 93.40$\pm$0.25 & 93.15$\pm$0.47 & 87.25$\pm$0.73 & 89.02$\pm$1.08 & 85.12$\pm$1.62 & 81.95$\pm$2.56 & 93.36$\pm$0.89 & 90.05$\pm$0.84 \\ \hline
LwTR (DA) & 94.11$\pm$0.68 & 96.33$\pm$0.22 & 95.36$\pm$0.29 & 97.79$\pm$0.31 & \textbf{94.11$\pm$0.27\textsuperscript{†}} & 92.76$\pm$0.25 & 86.95$\pm$0.61 & 89.74$\pm$0.72 & 86.20$\pm$1.67 & 83.08$\pm$1.78 & 93.70$\pm$0.64 & 90.28$\pm$0.55 \\
MR (DA)   & 93.43$\pm$0.13 & 96.18$\pm$0.33 & 95.01$\pm$0.69 & 97.69$\pm$0.12 & 93.15$\pm$0.60 & 92.67$\pm$0.32 & 86.78$\pm$1.12 & \textbf{90.06$\pm$0.61} & 86.36$\pm$1.64 & \textbf{83.81$\pm$2.79} & \textbf{93.69$\pm$0.61} & \textbf{90.69$\pm$1.23} \\
SR (DA)   & 94.18$\pm$0.92 & 96.21$\pm$0.30 & 95.45$\pm$0.30 & 97.87$\pm$0.14 & 93.41$\pm$0.23 & 93.39$\pm$0.29 & 86.78$\pm$1.49 & 89.61$\pm$0.56 & 86.42$\pm$2.36 & 81.83$\pm$2.85 & 93.53$\pm$0.72 & 90.04$\pm$0.97 \\
MELM (DA) & 94.07$\pm$0.67 & 96.09$\pm$0.14 & \textbf{95.67$\pm$0.71} & \textbf{97.83$\pm$0.12} & 92.84$\pm$0.73 & \textbf{93.43$\pm$0.64} & 86.38$\pm$1.16 & 89.05$\pm$1.18 & 86.65$\pm$1.37 & 81.89$\pm$2.77 & 93.30$\pm$0.59 & 89.12$\pm$1.47 \\ \hline
\end{tabular}
}
\caption{AUPRC scores of existing methods and data augmentation methods in OntoNotes 5.0.}
\label{tab:ontonotes5_auprc_scores}
\end{table*}


The results shown in Table \ref{tab:ontonotes5_id_scores} demonstrate ECE and MCE in OntoNotes 5.0 for NER in the ID setting, which the source domain and target domain are the same. The table results show that data augmentation methods consistently have better calibration performance than TS, LS, and MC Dropout, which have been considered to work for general classification problems, in the evaluation of calibration performance, in the ID setting. In particular, when the source genre is $\mathtt{tc}$, MELM and other data augmentation methods show superior calibration performance, with up to 6.01 $\%$ improvement for ECE and 5.62 $\%$ improvement for MCE compared to Baseline. As shown in Table \ref{tab:datasets}, the $\mathtt{tc}$ domain is not a data-poor setting, where there is sufficient training data and data augmentation is generally effective. MR and SR also show good calibration performance following MELM. Moreover, we can see that applying data augmentation methods do not increase inference time (See Appendix~\ref{sec:appendix_inference_time} Table~\ref{tab:inference_time}).
On the other hand, as Table \ref{tab:ontonotes5_ood_scores} shows, when the target domain is OOD, especially when the target (e.g. OntoNotes 5.0 $\mathtt{wb}$) is far from the source domain, the degree of improvement in the uncertainty estimation performance of data augmentation is not large, and sometimes even decreases.

\begin{figure}[t!]
    \centering
    \includegraphics[width=0.40\textwidth]{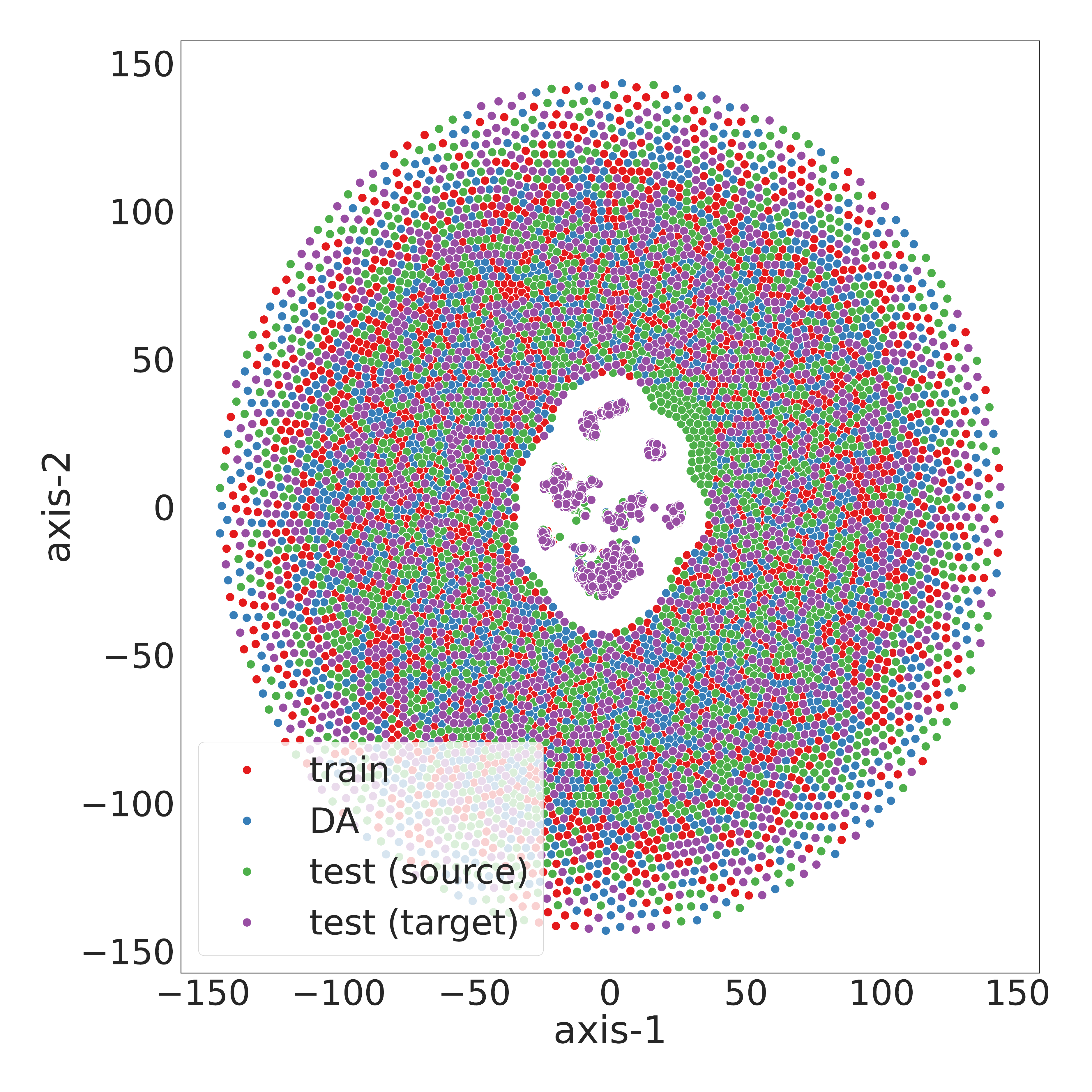}
    \caption{t-SNE plot of token embeddings of OntoNotes 5.0 $\mathtt{bn}$ training set (red), generated data by MELM (blue), source domain test set (green) and OntoNotes 5.0 $\mathtt{wb}$ test set (purple), respectively.}
    \label{fig:tsne_token_embeddings}
\end{figure}

We presume that the augmented data is not far from the original training set, because data augmentation methods we targeted in this study are based on the replacement of tokens or entities. Considering a recent study that indicates models tend to be more overconfident in areas with less training data~\cite{xiong2023proximityinformed}, we can consider calibration performance in OOD sets, especially far from the source domain, will not improve by data augmentation for NER, while the performance in ID sets will be better than existing methods.

To illustrate this, we performed t-SNE~\cite{JMLR:v9:tsne_vandermaaten08a} for the token embeddings with only entity token from trained Baseline model, shown in Figure \ref{fig:tsne_token_embeddings}. We can understand that the token embeddings from augmented data are near the train set or ID test set, while the OOD test sets have some poorly covered regions. Generating sentences that are distant from the training data set and semantically aligned entities from label description for uncertainty estimation is an interesting direction for future research.

AUPRC scores are shown in Table \ref{tab:ontonotes5_auprc_scores}. In the AUPRC scores in OntoNotes 5.0, data augmentation methods are outperform existing methods in 15 cases out of 24 cases. Among the existing methods, TS shows superior performance; in data augmentation methods, MELM is not as good as in the case of calibration metrics such as ECE and MCE, and MR tends to show superior uncertainty performance. Calibration and scores based on AUC measure different points of uncertainty~\cite{galil2023what}, therefore we assume that uncertainties that can be improved vary depending on the methods.

\subsection{Cross-lingual Evaluation}
\label{sec:cross-lingual-calibration}

\begin{table*}[t]
\centering
\scalebox{0.53}{
\begin{tabular}{l|ccc|ccc|ccc|ccc}
\hline
\multicolumn{1}{l|}{Methods} & \multicolumn{12}{c}{MultiCoNER ($\mathtt{EN}$)} \\ \cline{2-13}
\multicolumn{1}{l|}{} & \multicolumn{3}{c|}{$\mathtt{EN}$} & \multicolumn{3}{c|}{$\mathtt{DE}$} & \multicolumn{3}{c|}{$\mathtt{ES}$} & \multicolumn{3}{c}{$\mathtt{HI}$} \\ \cline{2-13}
\multicolumn{1}{c|}{} & ECE (↓) & MCE (↓) & AUPRC (↑) & ECE (↓) & MCE (↓) & AUPRC (↑) & ECE (↓) & MCE (↓) & AUPRC (↑) & ECE (↓) & MCE (↓) & AUPRC (↑) \\ \hline
Baseline   & 28.29$\pm$0.30  & 30.51$\pm$0.39 & 93.04$\pm$0.18 & 31.31$\pm$0.52 & 34.91$\pm$0.83 & 91.97$\pm$0.23 & 31.22$\pm$0.28 & 33.70$\pm$0.39  & 90.87$\pm$0.27 & 46.84$\pm$1.64 & 48.13$\pm$1.51 & 82.04$\pm$2.24 \\
TS         & 28.46$\pm$0.43 & 30.70$\pm$0.52  & 93.13$\pm$0.17 & 31.45$\pm$0.70  & 35.08$\pm$1.05 & 92.02$\pm$0.24 & 31.24$\pm$0.41 & 33.77$\pm$0.38 & 90.92$\pm$0.18 & 46.83$\pm$1.38 & 48.35$\pm$1.25 & 83.01$\pm$1.45 \\
LS         & 28.50$\pm$0.57  & 30.60$\pm$0.68  & 93.12$\pm$0.13 & 31.50$\pm$0.64 & 34.81$\pm$0.66 & 91.93$\pm$0.26 & 31.43$\pm$0.58 & 33.83$\pm$0.67 & 90.82$\pm$0.10 & 46.36$\pm$1.23 & 47.95$\pm$1.03 & 84.00$\pm$1.60   \\
MC Dropout  & 28.57$\pm$0.34 & 30.83$\pm$0.54 & 92.97$\pm$0.34 & 31.64$\pm$0.48 & 35.24$\pm$0.68 & 91.86$\pm$0.37 & 31.47$\pm$0.42 & 33.98$\pm$0.40 & 90.79$\pm$0.22 & 47.42$\pm$1.30 & 48.77$\pm$1.23 & 81.39$\pm$3.30  \\ \hline
LwTR (DA)  & 28.17$\pm$0.54 & 30.48$\pm$0.77 & 92.80$\pm$0.28  & 31.13$\pm$0.59 & 34.60$\pm$0.78  & 91.57$\pm$0.34 & 31.10$\pm$0.35  & 33.61$\pm$0.51 & 90.66$\pm$0.27 & 46.70$\pm$1.47  & 47.95$\pm$1.30  & 82.57$\pm$1.96 \\
MR (DA)    & \textbf{28.01$\pm$0.42}  & \textbf{30.08$\pm$0.49\textsuperscript{†}} & \textbf{93.30$\pm$0.24}  & \textbf{31.12$\pm$0.74} & 34.71$\pm$0.81 & \textbf{92.05$\pm$0.20}  & \textbf{30.75$\pm$0.34\textsuperscript{†}} & \textbf{33.24$\pm$0.36\textsuperscript{†}} & \textbf{91.03$\pm$0.15} & 46.96$\pm$1.20  & 48.28$\pm$1.12 & 81.75$\pm$2.52 \\
SR (DA)    & 28.15$\pm$0.42 & 30.36$\pm$0.48 & 93.08$\pm$0.26 & 31.17$\pm$0.39 & \textbf{34.42$\pm$0.70}  & 92.02$\pm$0.39 & 31.60$\pm$0.55  & 33.86$\pm$0.56 & 90.65$\pm$0.33 & \textbf{45.85$\pm$0.53} & \textbf{47.38$\pm$0.47} & \textbf{84.91$\pm$0.91} \\
MELM (DA)  & 28.53$\pm$0.38 & 30.68$\pm$0.43 & 92.72$\pm$0.22 & 32.61$\pm$0.49 & 36.14$\pm$0.65 & 91.17$\pm$0.29 & 32.09$\pm$0.44 & 34.38$\pm$0.52 & 90.14$\pm$0.30  & 47.91$\pm$1.79 & 49.18$\pm$1.79 & 81.13$\pm$2.41 \\ \hline
\end{tabular}
}
\caption{Results of existing calibration methods and data augmentation methods in MultiCoNER.}
\label{tab:multiconer_EN_calibration_scores}
\end{table*}

The results of cross-lingual transfer in MultiCoNER are shown in Table \ref{tab:multiconer_EN_calibration_scores} with English as the source language. MR performs better in uncertainty performance for the ID situation. In contrast to the calibration and uncertainty performance in the cross-genre setting, both MR and SR show better calibration and uncertainty in the OOD setting. In \citet{jiang-etal-2022-calibrating}, the result shows that the larger the linguistic distance \cite{linguistic-distance-2005}, the more lenient the calibration and uncertainty estimation tends to be, and similar trends are obtained in this experiment. Unlike the discussion in Section \ref{sec:cross-genre-calibration},  the uncertainty performance by data augmentation is also good for OOD in cross-lingual setting because the areas where only target set exist is limited in MultiCoNER (illustrated in Appendix \ref{sec:appendix_tsne_multiconer}).  On the other hand, MELM, which tends to show excellent calibration performance in cross-genre calibration, does not show good performance in cross-lingual settings.

The amount of data for each language in the CC100 \cite{conneau-etal-2020-unsupervised} dataset used to train the base model, mDeBERTaV3, was highest for English, followed by German, Spanish, Hindi, and Bangla which correlates with the trend of the calibration results. Moreover, as mentioned in \citet{limisiewicz-etal-2023-tokenization}, languages that tend to have vocabulary overlap between languages in tokenization perform better in cross-lingual transfer in NER. Similar effects may be observed in confidence calibration and uncertainty estimation.

\subsection{Detailed Analyzes}
\label{sec:detailed-analysis}
We investigate the effects of entity overlap rates and the perplexity of the generated sentences to gain a better understanding of the confidence calibration and uncertainty estimation performance of data augmentation methods for NER. We also investigate the impact of data augmentation size in several settings.

\begin{figure*}[t!]
    \centering
    \includegraphics[width=0.95\textwidth]{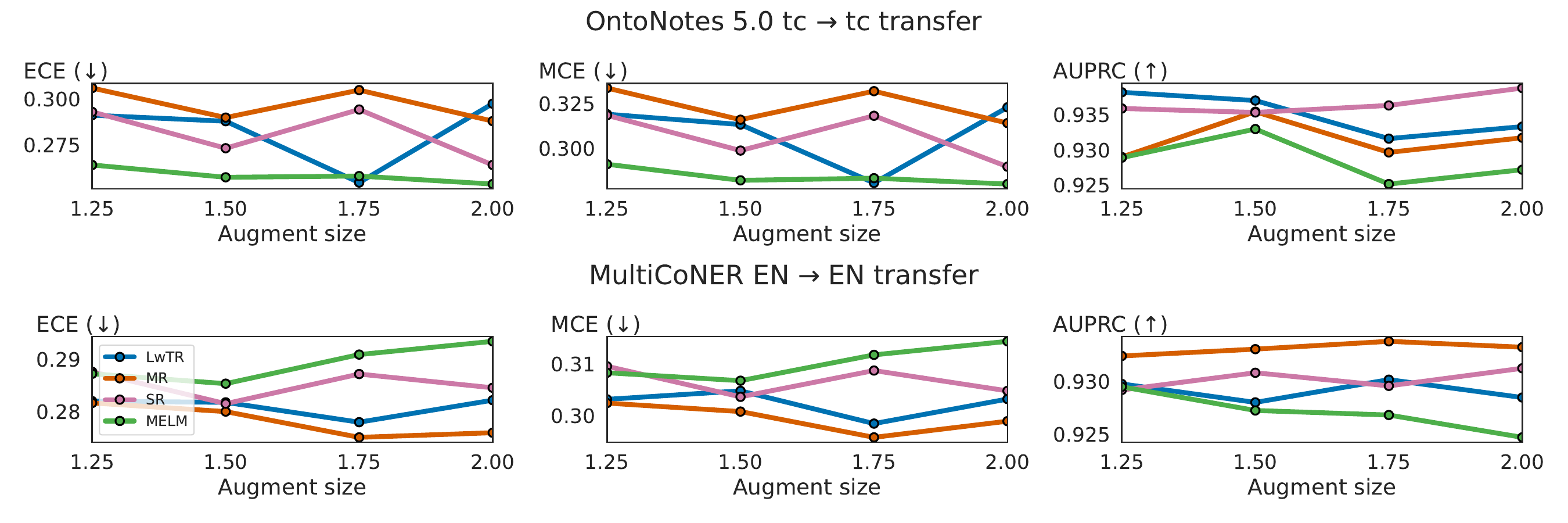}
    \caption{Average values of evaluation metrics for each data augmentation method in ID settings.}
    \label{fig:id_transfer_augmentation_size}
\end{figure*}

\begin{figure*}[t!]
    \centering
    \includegraphics[width=1.0\textwidth]{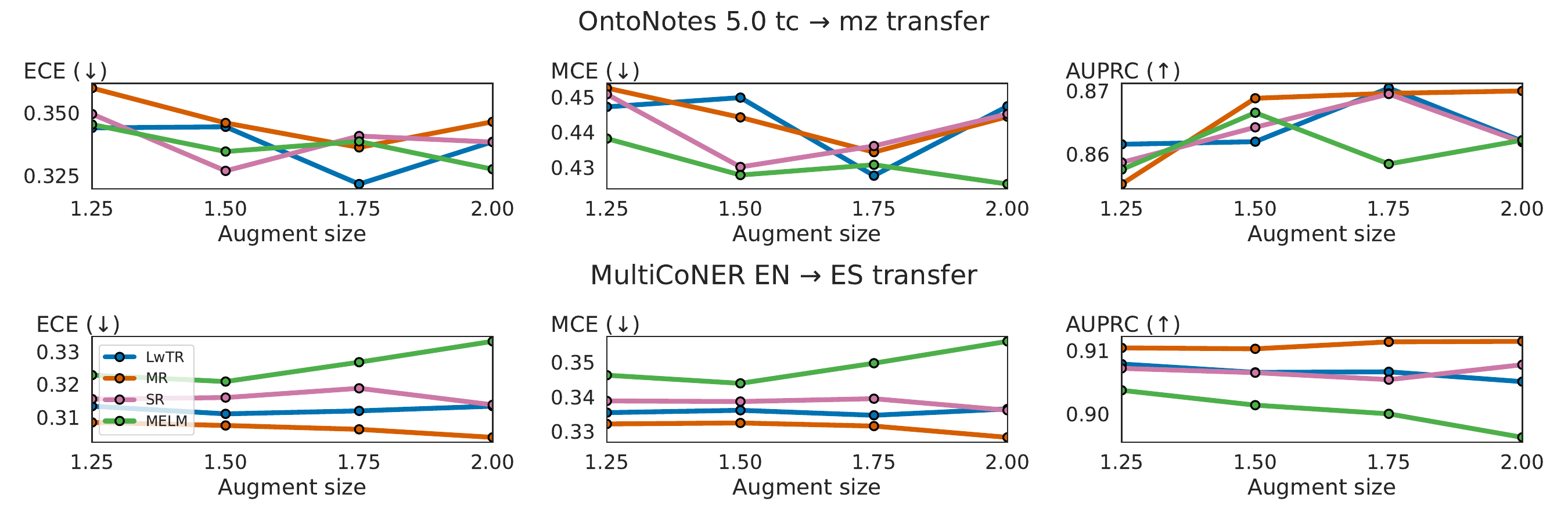}
    \caption{Average values of evaluation metrics for each data augmentation method in OOD settings.}
    \label{fig:ood_transfer_augmentation_size}
\end{figure*}

\begin{table*}[t!]
\centering
\scalebox{0.80}{
\begin{tabular}{lccccc}
\hline
Methods & OntoNotes 5.0 ($\mathtt{bc}$) & OntoNotes 5.0 ($\mathtt{bn}$) & OntoNotes 5.0 ($\mathtt{nw}$) & OntoNotes 5.0 ($\mathtt{tc}$) & MultiCoNER ($\mathtt{EN}$) \\
\hline 
LwTR & 7.05 & 7.59 & 8.28 & 7.33 & 6.78 \\
MR & \textbf{5.36} & \textbf{5.27} & \textbf{5.27} & \textbf{5.83} & \textbf{5.83} \\
SR & 5.91 & 6.35 & 6.62 & 6.02 & 6.35 \\
MELM & 5.56 & 5.65 & 5.55 & 5.90  & 6.14 \\ \hline
(Train)  & 5.18 & 4.84 & 4.86 & 5.80 & 5.54 \\
\hline
\end{tabular}
}
\caption{Sentences perplexities generated by the data augmentation method in each dataset. Each data augmentation method is performed to increase the training data. Bold means the lowest score in data augmentation methods.}

\label{tab:perplexities}
\end{table*}

\subsubsection{Impact of Augmentation Size}
\label{sec:impact-of-augmentation-size}
To investigate the impact of data augmentation size on calibration and uncertainty performance, we analyze the trend of evaluation metrics in $\mathtt{tc}$ → $\mathtt{mz}$ scenario of OntoNotes 5.0 and $\mathtt{EN}$ → $\mathtt{ES}$ scenario of MultiCoNER, respectively. Figure \ref{fig:id_transfer_augmentation_size} and \ref{fig:ood_transfer_augmentation_size} illustrate the results in the ID and OOD settings, respectively. In many cases, MR improves the calibration and uncertainty performance by increasing data.\footnote{Note that we have not discussed about the absolute values of the uncertainty estimation performance.} SR consistently improves as the dataset size doubles, whereas LwTR demonstrates only marginal improvement or even worsens as the dataset size increases. Finally, MELM improves further for OntoNotes 5.0 $\mathtt{tc}$, which shows excellent performance, and deteriorates further for MultiCoNER $\mathtt{EN}$, which shows poor performance.

These results show that the calibration algorithm with the best performance for cross-domain transfers is likely to have better performance as the augmentation size is increased. On the other hand, increasing the augmentation size in MR improves the calibration and uncertainty performance compared to similar other data augmentation methods. \par
Since data augmentation by MR and MELM is performed only on the entity region, the uncertainty estimation performance is relatively less adversely affected by increasing the data augmentation size. On the other hand, in SR and LwTR, data augmentation that replaces tokens may often inject tokens with inappropriate parts of speech for that sentence, so increasing the data augmentation size often leads to a degradation of uncertainty estimation performance.

\subsubsection{Impact of Perplexities for Augmented Sentences}
\label{sec:impact-of-perplexities}
To investigate the influence of replacement units on data augmentation for NER as mentioned in Section \ref{sec:impact-of-augmentation-size}, we measured the perplexity of the augmented sentences using GPT-2 \cite{radford2019language}. The average perplexities of the augmented sentences and the average perplexities of the original training set for each dataset are shown in Table \ref{tab:perplexities}. Lower perplexity from augmented sentences tends to improve calibration performance and uncertainty performance.
Consistently, the average perplexity of the sentences generated by MR is the lowest. Since MR performs substitutions on an entity-by-entity basis and does not affect the structure of the sentence itself, it has the lowest perplexity among the data augmentation methods in NER.\footnote{As shown in Appendix~\ref{sec:appendix_f1_scores}, not only the uncertainty performance but also the prediction performance could be affected by preserving the structure of a sentence.} MELM has the second lowest perplexity after MR, and may be adversely affected by generated entities that are adapted to the context but not actually present.

\section{Conclusion}
In this paper, we investigated the impact of data augmentation on the confidence calibration and uncertainty estimation in NER in terms of genre and language, using several metrics. First, we find that MELM, MR, and SR lead to better calibration and uncertainty performance in the ID setting consistently. On the other hand, in the OOD setting, uncertainty estimation by data augmentation is less effective, especially when the target domain is far from the source domain. Second, our results suggest that the lower the perplexity of the augmented data, as in MR, the further better the calibration and uncertainty performance as the augmentation size is increased. Data augmentation methods for NER do not require changes to the model structure and only require more data to improve entity-level calibration and performance without the need to change the model structure. Our findings indicate the effectiveness of uncertainty estimation through data augmentation for NER, and will be expected to stimulate future research based on their limitations.

\section*{Limitations}
While this experiment provided valuable insights into the impact of data augmentation on confidence calibration and uncertainty estimation in NER across different genres and languages, there are several limitations that should be acknowledged. \par

\paragraph{Source Language} Due to resource limitations, the experiment was limited to evaluation with English as the source language. To effectively investigate the calibration and uncertainty of zero-shot cross-lingual transfer, it is important to expand the investigation to include a wider range of languages as the source language. Therefore, future research should prioritize the investigation of calibration and uncertainty performance using different languages as the source for zero-shot cross-lingual transfer. \par

\paragraph{Evaluation of Uncertainty for Entities} As mentioned in Section \ref{sec:eval-details}, regarding the calibration and uncertainty evaluation policy, we simply evaluated an entity span as a single data instance, but a rigorous evaluation method that performs evaluation while considering multiple span candidates has been proposed \cite{jiang-etal-2022-calibrating}. Establishing span-level NER calibration evaluation methods that can efficiently and comprehensively evaluate calibration and uncertainty for entity types for datasets with many entity types and long sequence lengths is a topic for future research. \par

\paragraph{NER Paradigm} We broadly evaluated the calibration and uncertainty performance in both cross-genre and cross-lingual settings on data augmentation for NER, but only using sequence labeling-based methods. Recently, other paradigms in NER have been proposed such as the span-based methods \cite{fu-etal-2021-spanner} and the generation-based methods \cite{yan-etal-2021-unified-generative} including BART~\cite{lewis-etal-2020-bart} or Large Language Models (LLM) ~\cite{xu2024largelanguagemodelsgenerative}, which are also applicable to nested-NER. In the future, the calibration or uncertainty performance of these methods could be evaluated.

\paragraph{Other Data Augmentation Methods} In this study, we focused on the data augmentation methods based on token or entity replacement. On the other hand, paraphrase-based data augmentation methods using such as LLM have attracted attention~\cite{ding2024dataaugmentationusinglarge}. By using LLM, it is also possible to generate entities that correspond to a specified entity type~\cite{ye2024llmdadataaugmentationlarge}. To Investigate these in the context of uncertainty estimation also will be an interesting research.

\section*{Ethical Considerations}
In this study, we used existing datasets that have cleared ethical issues. Furthermore, the data augmentation methods we used for uncertainty estimation are substitution-based methods except for MELM, and MELM generated entities from existing datasets that have no ethical issues. Therefore, it is unlikely that toxic sentences would be generated.

\section*{Acknowledgements}
The authors also acknowledge the Nara Institute of Science and Technology's HPC resources made available for conducting the research reported in this paper.

\bibliography{custom}

\appendix

\section{Licenses of Datasets}
\label{sec:appendix_licenses_datasets}
OntoNotes 5.0 can be used for research purposes as described in \url{https://catalog.ldc.upenn.edu/LDC2013T19}. MultiCoNER dataset is licensed by CC BY 4.0 as described in \url{https://aws.amazon.com/marketplace/pp/prodview-cdhrtt7vq4hf4}.

\section{Details of Existing Calibration Methods}
\label{sec:appendix_details_existing_baseline_methods}
In this section, we describe the popular baseline methods for confidence calibration. We use the following notations: $\bm{z}_i$ denotes the logits for class $i$, $p_i$ denotes the calibrated probability for class $i$, $y_i$ denotes the label for class $i$, and $K$ denotes the number of classes.

\subsection{Temperature Scaling (TS)}
TS \cite{pmlr-v70-guo17a} is a post-processing technique for calibrating the confidence scores outputted by a neural network. It involves scaling the logits (i.e., the outputs of the final layer before the softmax) by a temperature parameter $T$ before applying the softmax function to obtain the calibrated probabilities. The softmax function takes a vector of logits $\boldsymbol{z}$ and returns a distribution $\boldsymbol{p}$:

\begin{equation*}
p_i = \frac{\exp(\bm{z}_i/T)}{\sum_{j=1}^K \exp(\bm{z}_j/T)} .
\end{equation*}

\subsection{Label Smoothing (LS)}
LS \cite{Miller1996AGO, pereyra2017regularizing} is a regularization technique used to improve the calibration and generalization performance of the model. By introducing a small degree of uncertainty in the target labels during training, label smoothing mitigates overfitting and encourages the model to learn more robust and accurate representations, ultimately contributing to improved overall performance on the task at hand. LS is characterized by introducing a smoothing parameter $\epsilon$ and smoothed label $y^{LS}_{i}$ as follows,

\begin{equation*}
y^{LS}_{i} = y_{i} (1 - \epsilon) + \frac{\epsilon}{K}.
\end{equation*}

\begin{table}[t!]
\centering
\scalebox{0.80}{
\begin{tabular}{lr}
\hline
Methods & Inference time [s] \\\hline
Baseline & 14.90 $\pm$ 0.10 \\
TS & 15.53 $\pm$ 0.92 \\
LS & 14.94 $\pm$ 0.24 \\
MC Dropout & 271.77 $\pm$ 1.81 \\
LwTR & 14.91 $\pm$ 0.10 \\
MR & 14.93 $\pm$ 0.17 \\
SR & 14.83 $\pm$ 0.12 \\
MELM & 14.89 $\pm$ 0.14 \\
\hline
\end{tabular}
}
\caption{Inference time for each algorithm on MultiCoNER $\mathtt{EN}$ full test data.}
\label{tab:inference_time}
\end{table}

\subsection{Monte-Carlo Dropout (MC Dropout)}
MC Dropout is a regularization technique that can be used for uncertainty estimation in neural networks \cite{pmlr-v48-gal16}. In this method, we need to run the model $M$ times with different dropout masks and take the average softmax output over all the runs (We use $M = 20$). The procedure can be represented using the following formula:
\begin{equation*}
p_i = \frac{1}{M} \sum_{t=1}^M \frac{\exp(\bm{z}_i^{(t)})}{\sum_{j=1}^K \exp(\bm{z}_j^{(t)})}. 
\end{equation*}

\begin{table*}[t!]
\centering
\scalebox{0.62}{
\begin{tabular}{l|cc|cc|cc|cc}
\hline
\multicolumn{1}{l|}{Methods} & \multicolumn{2}{c|}{$\mathtt{bc}$} & 
\multicolumn{2}{c|}{$\mathtt{bn}$} & \multicolumn{2}{c|}{$\mathtt{nw}$} & \multicolumn{2}{c}{$\mathtt{tc}$} \\ \cline{2-9} 
\multicolumn{1}{c|}{} & ECE (↓) & MCE (↓) & ECE (↓) & MCE (↓) & ECE (↓) & MCE (↓) & ECE (↓) & MCE (↓) \\ \hline
Baseline & \textbf{27.07} & 33.52 & 26.08 & 31.17 & 26.66 & 31.35 & 37.66 & 46.32 \\
TS  & 27.25 & 33.41 & 26.17 & 31.17 & 26.68 & 31.34 & 36.66 & 45.52 \\
LS & 27.19 & 33.57 & \textbf{25.88} & \textbf{30.49} & 26.52 & 30.67 & 35.24 & 43.68 \\
MC Dropout & 27.15 & 33.61 & 25.90 & 30.85 & 26.62 & 31.18 & 36.80 & 45.71 \\ \hline
LwTR (DA) & 27.65 & 33.78 & 26.49 & 31.78 & 27.28 & 31.67 & 35.90 & 44.97 \\
MR (DA) & 27.33	& 33.22 & 26.21 & 31.00 & \textbf{26.26} & \textbf{30.53} & 36.38	& 44.65 \\
SR (DA) & 27.23	& \textbf{33.08} & 26.11 & 30.72 & 27.47 & 31.89 & 35.24 & 43.57 \\
MELM (DA) & 27.95 & 33.88 & 26.63 & 30.91 & 27.62 & 32.09 & \textbf{34.83} & \textbf{42.65} \\ \hline
\end{tabular}
}
\caption{ECE and MCE averaged over all target domain results in OntoNotes 5.0.}
\label{tab:ontonotes5_ece_mce_averaged_score}
\end{table*}

\section{Inference Time}
\label{sec:appendix_inference_time}

Table~\ref{tab:inference_time} shows the results of the inference time on MultiCoNER $\mathtt{EN}$ set. We can see that data augmentation methods do not affect the computational overhead during inference clearly.

\section{Full Averaged Results on OntoNotes 5.0}
\label{sec:appendix_full_averaged_ontonotes}
To briefly summarize the many values in Table~\ref{tab:ontonotes5_id_scores} and ~\ref{tab:ontonotes5_ood_scores}, we averaged the ECE and MCE scores for each method and domain, shown them in Table~\ref{tab:ontonotes5_ece_mce_averaged_score}. From this table, we can see that data augmentation methods are slightly worse than existing methods some cases when averaging all settings , while in others, especially $\mathtt{nw}$ and $\mathtt{tc}$, data augmentation methods are better on average.

\section{More Results about Test Set Duplication}
\label{sec:appendix_other_results_testset_duplication}
Table \ref{tab:appendix_duplication_test_set_percentages_others} shows the results of the percentage increase in entity duplication that are new overlaps with each target domain’s test set when applying each data augmentation method except MR, where the source domains are $\mathtt{bc}$, $\mathtt{bn}$, and $\mathtt{nw}$. In all cases there is only a small increase. These results and the MR, which shows good calibration and uncertainty performance indicated from Section \ref{sec:cross-genre-calibration} and \ref{sec:cross-lingual-calibration}, do not increase the number of new entities in the training data set suggest that the entity overlap rate does not affect calibration and uncertainty estimation.

\begin{table}[t!]
\centering
\scalebox{0.88}{
\begin{tabular}{l|cccccc}
\hline
\multicolumn{1}{l|}{Methods} & \multicolumn{6}{c}{OntoNotes 5.0 ($\mathtt{bc}$)} \\ \cline{2-7}
\multicolumn{1}{c|}{} & $\mathtt{bc}$ & $\mathtt{bn}$ & $\mathtt{mz}$ & $\mathtt{nw}$ & $\mathtt{tc}$ & $\mathtt{wb}$\\
\hline 
LwTR  & 0.27 & 0.26 & 0.00 & 0.14 & 1.83 & 0.30 \\
SR    & 0.00 & 0.18 & 0.00 & 0.14 & 0.00 & 0.15 \\
MELM  & 0.41 & 0.53 & 0.19 & 0.17 & 0.91 & 0.45 \\ \hline \hline
\multicolumn{1}{l|}{Methods} & \multicolumn{6}{c}{OntoNotes 5.0 ($\mathtt{bn}$)} \\ \cline{2-7}
\multicolumn{1}{c|}{} & $\mathtt{bc}$ & $\mathtt{bn}$ & $\mathtt{mz}$ & $\mathtt{nw}$ & $\mathtt{tc}$ & $\mathtt{wb}$\\
\hline 
LwTR  & 0.55 & 0.35 & 0.19 & 0.35 & 0.91 & 0.60 \\
SR    & 0.55 & 0.26 & 0.19 & 0.21 & 0.00 & 0.45 \\
MELM  & 0.68 & 0.35 & 0.37 & 0.10 & 0.46 & 0.30 \\ \hline \hline
\multicolumn{1}{l|}{Methods} & \multicolumn{6}{c}{OntoNotes 5.0 ($\mathtt{nw}$)} \\ \cline{2-7}
\multicolumn{1}{c|}{} & $\mathtt{bc}$ & $\mathtt{bn}$ & $\mathtt{mz}$ & $\mathtt{nw}$ & $\mathtt{tc}$ & $\mathtt{wb}$\\
\hline 
LwTR  & 0.96 & 1.23 & 0.37 & 0.52 & 5.02 & 1.34 \\
SR    & 0.41 & 0.09 & 0.56 & 0.21 & 0.46 & 1.04 \\
MELM  & 1.10 & 0.79 & 1.48 & 0.55 & 1.37 & 0.45 \\ \hline
\end{tabular}
}

\caption{The percentage of new entities generated by each data augmentation method using the training set in the case of the source domain $\mathtt{bc}$, $\mathtt{bn}$ and  $\mathtt{nw}$.}
\label{tab:appendix_duplication_test_set_percentages_others}
\end{table}

\begin{table*}[t!]
\centering
\scalebox{0.80}{
\begin{tabular}{lccccc}
\hline
Methods & OntoNotes 5.0 ($\mathtt{bc}$) & OntoNotes 5.0 ($\mathtt{bn}$) & OntoNotes 5.0 ($\mathtt{nw}$) & OntoNotes 5.0 ($\mathtt{tc}$) & MultiCoNER ($\mathtt{EN}$) \\
\hline 
LwTR &27.77  & 32.69  & 38.65 & 19.83  & 18.46 \\
MR & 0.00 & 0.00 & 0.00 & 0.00  & 0.00 \\
SR & 25.23 & 26.34 & 35.13 & 8.56 & 20.45\\
MELM & 45.26 & 45.95 & 43.37 & 34.75 & 37.64 \\ \hline
\end{tabular}
}

\caption{The percentage increase in new entities when each data augmentation method is performed on the original train set.}
\label{tab:new_entities_rate}
\end{table*}

\begin{table}[t!]
\centering
\scalebox{0.85}{
\begin{tabular}{lcccccc}
\hline
Methods & $\mathtt{bc}$ & $\mathtt{bn}$ & $\mathtt{mz}$ & $\mathtt{nw}$ & $\mathtt{tc}$ & $\mathtt{wb}$\\
\hline 
LwTR & 0.00 & 0.00 & 0.00 & 0.10 & 0.00 & 0.00 \\
SR & 0.14 & 0.00 & 0.00 & 0.10 & 0.00 & 0.15 \\
MELM & 0.27 & 0.35 & 0.19 & 0.14 & 0.00 & 0.30 \\ \hline
\end{tabular}
}

\caption{The percentage increase in entity duplication in the case of the source domain $\mathtt{tc}$ that are new overlaps with each target domain’s test set when applying each data augmentation method except MR. More results are in Appendix \ref{sec:appendix_other_results_testset_duplication}.}
\label{tab:duplication_test_set_percentages}
\end{table}

\section{Impact of New Entities via Data Augmentation}
\label{sec:impact-of-entities-duplication}
To investigate the impact of new entities added by data augmentation methods on calibration performance, we measured the percentage of new entities added in the training data and the percentage of new entities that overlap with the test set. Table \ref{tab:new_entities_rate} shows the percentage of new entities increased by data augmentation with the train set as the source domain in each dataset. In all data sets, MELM has observed the most increase of the new entities in the augmented data set. On the other hand, MR that shows good calibration performance followed by MELM does not increase the number of new entities because the replacement is based on the entities in the original training data. Furthermore, the entities generated have little overlap with the target domain, as shown in Table \ref{tab:duplication_test_set_percentages}. Therefore, new entities by data augmentation methods for NER are likely to have no effect on calibration performance or uncertainty performance.

\section{t-SNE Plot for MultiCoNER Dataset}
\label{sec:appendix_tsne_multiconer}
\begin{figure}[t]
    \centering
    \includegraphics[width=0.40\textwidth]{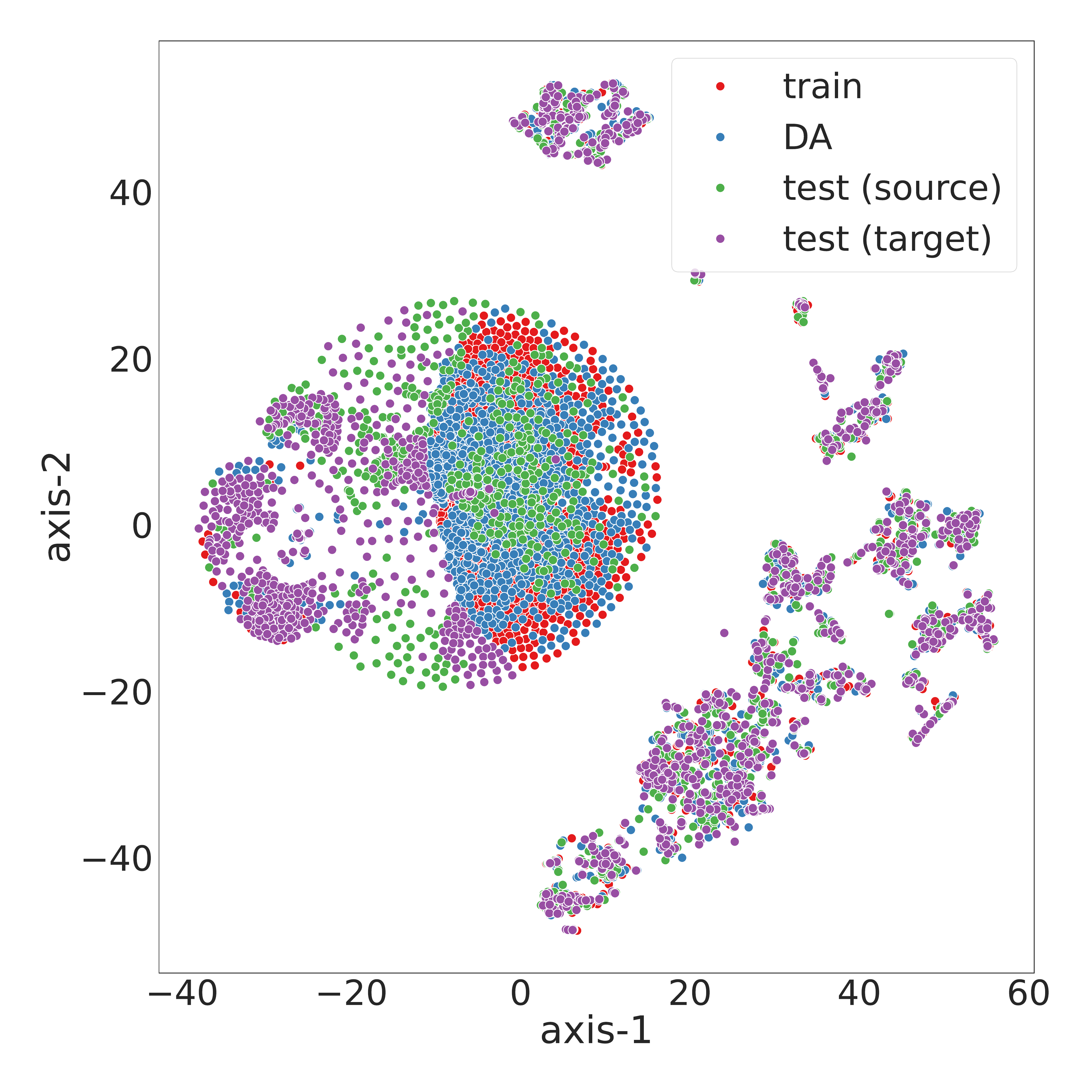}
    \caption{t-SNE plot of token embeddings of MultiCoNER $\mathtt{EN}$ training set (red), generated data by SR (blue), source domain test set (green) and MultiCoNER $\mathtt{HI}$ test set (purple), respectively.}
    \label{fig:tsne_token_embeddings_multiconer}
\end{figure}

To overview of the ID and OOD data instances in the MultiCoNER dataset, t-SNE plot is shown in Figure \ref{fig:tsne_token_embeddings_multiconer}.

\section{Results for Low-resource Language}
\label{sec:appendix_multiconer_lowresource}
To investigate the uncertainty estimation performance for low-resource language, we additionaly show the results of 10,000 examples of Bangla ($\mathtt{BN}$) from MultiCoNER dataset in Table~\ref{tab:multiconer_BN_scores} when source language is $\mathtt{EN}$. The results show that data augmentation is also effective in uncertainty estimation for low-resource language.

\begin{table}[t!]
\centering
\scalebox{0.65}{
\begin{tabular}{l|ccc}
\hline
\multicolumn{1}{l|}{Methods} & ECE (↓) & MCE (↓) & AUPRC (↑) \\ \hline
Baseline   & 49.60$\pm$2.02  & 51.32$\pm$1.96 & 79.49$\pm$2.21 \\
TS         & 48.85$\pm$1.89	 & 50.60$\pm$1.60  & 79.09$\pm$4.22 \\
LS         & 48.00$\pm$1.97  & 49.91$\pm$1.54  & 79.60$\pm$3.51 \\
MC Dropout  & 49.29$\pm$2.20 & 50.93$\pm$2.14 & 78.31$\pm$2.52 \\ \hline
LwTR (DA)  & 48.66$\pm$1.35		 & 50.22$\pm$1.36 & 80.93$\pm$1.75 \\
MR (DA)    & 49.54$\pm$2.65  & 51.20$\pm$2.65 & 79.17$\pm$2.97 \\
SR (DA)    & \textbf{47.67$\pm$0.98} & \textbf{49.46$\pm$0.88} & \textbf{81.96$\pm$1.35} \\
MELM (DA)  & 50.77$\pm$0.88 & 52.15$\pm$0.81 & 75.55$\pm$2.59 \\ \hline
\end{tabular}
}
\caption{Results of existing methods and data augmentation methods in MultiCoNER $\mathtt{BN}$.}
\label{tab:multiconer_BN_scores}
\end{table}

\begin{table*}[t!]
\centering
\scalebox{0.53}{
\begin{tabular}{l|c|c|c|c|c|c||c|c|c|c|c|c}
\hline
\multicolumn{1}{l|}{Methods} & \multicolumn{6}{c||}{OntoNotes 5.0 ($\mathtt{bc}$)} 
 & \multicolumn{6}{c}{OntoNotes 5.0 ($\mathtt{bn}$)} \\ \cline{2-13}
\multicolumn{1}{c|}{} & $\mathtt{bc}$ & $\mathtt{bn}$ & $\mathtt{mz}$ & $\mathtt{nw}$ & $\mathtt{tc}$ & $\mathtt{wb}$ & $\mathtt{bc}$ & $\mathtt{bn}$ & $\mathtt{mz}$ & $\mathtt{nw}$ & $\mathtt{tc}$ & $\mathtt{wb}$ \\ \hline
Baseline   & 81.39$\pm$0.78 & 80.86$\pm$1.03 & 81.61$\pm$1.36 & 75.49$\pm$0.90 & 68.83$\pm$1.27 & 45.74$\pm$0.74 & 80.74$\pm$1.21 & 90.25$\pm$0.36 & 81.47$\pm$0.96 & 81.04$\pm$0.64 & 72.36$\pm$1.88 & 46.86$\pm$0.52 \\
TS         & 81.10$\pm$0.94 & 81.19$\pm$0.89 & 80.80$\pm$1.37 & 75.14$\pm$1.60 & 69.20$\pm$2.73 & 45.58$\pm$1.02 & 81.31$\pm$1.18 & \textbf{90.37$\pm$0.49} & 80.96$\pm$1.32 & \textbf{81.13$\pm$0.62} & 71.83$\pm$1.76 & 46.50$\pm$0.69 \\
LS         & 81.21$\pm$1.11 & 81.17$\pm$0.91 & 81.43$\pm$1.33 & 75.30$\pm$1.26 & 69.64$\pm$1.45 & 45.75$\pm$0.82 & \textbf{82.08$\pm$0.62} & 90.32$\pm$0.36 & 81.22$\pm$0.52 & 80.95$\pm$0.37 & 72.45$\pm$1.38 & 46.69$\pm$0.60 \\
MC Dropout  & 81.49$\pm$0.80 & 81.06$\pm$0.71 & 81.12$\pm$0.63 & 75.24$\pm$1.02 & 69.53$\pm$1.78 & 45.73$\pm$0.46 & 81.55$\pm$0.63 & 90.21$\pm$0.36 & 80.80$\pm$1.10 & 81.11$\pm$0.46 & \textbf{73.13$\pm$1.97} & 46.71$\pm$0.60 \\ \hline
LwTR (DA)  & 80.85$\pm$0.82 & 80.91$\pm$0.93 & 81.45$\pm$1.08 & 75.33$\pm$0.82 & 68.40$\pm$0.94 & 45.53$\pm$0.84 & 79.43$\pm$1.13 & 89.98$\pm$0.40 & 80.75$\pm$0.67 & 80.33$\pm$0.31 & 69.62$\pm$1.80 & 46.23$\pm$0.54 \\
MR (DA)    & 80.93$\pm$0.61 & 80.88$\pm$0.61 & 
\textbf{82.02$\pm$0.66} & \textbf{75.66$\pm$0.79} & 69.49$\pm$1.78 & 45.38$\pm$0.72 & 79.93$\pm$1.43 & 90.07$\pm$0.23 & \textbf{81.70$\pm$0.61} & 80.54$\pm$0.50 & 72.44$\pm$1.46 & 46.45$\pm$0.47 \\
SR (DA)    & \textbf{81.52$\pm$0.69} & \textbf{81.20$\pm$0.78} & 79.93$\pm$0.95 & 75.08$\pm$0.89 & \textbf{69.86$\pm$1.30} & \textbf{46.04$\pm$0.57} & 80.24$\pm$1.44 & 90.05$\pm$0.21 & 80.92$\pm$0.93 & 80.84$\pm$0.42 & 70.80$\pm$1.66 & \textbf{46.98$\pm$0.61} \\
MELM (DA)  & 81.08$\pm$0.37 & 80.81$\pm$0.97 & 80.11$\pm$0.98 & 74.74$\pm$1.24 & 66.68$\pm$1.18 & 45.19$\pm$1.05 & 79.23$\pm$0.64 & 90.26$\pm$0.38 & 81.48$\pm$0.65 & 80.66$\pm$0.79 & 68.42$\pm$1.65 & 46.36$\pm$0.44 \\ \hline \hline
\multicolumn{1}{l|}{Methods} & \multicolumn{6}{c||}{OntoNotes 5.0 ($\mathtt{nw}$)} 
 & \multicolumn{6}{c}{OntoNotes 5.0 ($\mathtt{tc}$)} \\ \cline{2-13}
\multicolumn{1}{c|}{} & $\mathtt{bc}$ & $\mathtt{bn}$ & $\mathtt{mz}$ & $\mathtt{nw}$ & $\mathtt{tc}$ & $\mathtt{wb}$ & $\mathtt{bc}$ & $\mathtt{bn}$ & $\mathtt{mz}$ & $\mathtt{nw}$ & $\mathtt{tc}$ & $\mathtt{wb}$ \\ \hline
Baseline   & 74.34$\pm$4.10  & 83.08$\pm$1.19 & 73.56$\pm$3.31 & 90.08$\pm$0.31 & \textbf{72.59$\pm$1.34} & 46.47$\pm$0.59 & 55.29$\pm$2.01 & 59.13$\pm$2.80  & 50.68$\pm$3.51 & 46.14$\pm$4.31 & 69.52$\pm$1.45 & 40.85$\pm$1.36 \\
TS         & 75.34$\pm$1.67 & 83.02$\pm$0.98 & 75.01$\pm$2.21 & 90.04$\pm$0.24 & 71.98$\pm$1.17 & 46.29$\pm$0.87 & \textbf{56.81$\pm$2.05} & 59.04$\pm$2.95 & 52.98$\pm$3.34 & 48.85$\pm$3.26 & 67.45$\pm$2.30  & 41.12$\pm$1.27 \\
LS         & \textbf{76.60$\pm$1.65}  & \textbf{83.27$\pm$1.49} & \textbf{75.79$\pm$2.00}  & 90.20$\pm$0.26  & 71.91$\pm$2.67 & 46.68$\pm$0.69 & 53.98$\pm$3.40  & 56.12$\pm$6.02 & 51.17$\pm$5.94 & 48.62$\pm$4.82 & 66.01$\pm$3.26  & 40.63$\pm$1.83 \\
MC Dropout  & 75.07$\pm$2.84 & 82.69$\pm$2.11 & 73.79$\pm$2.23 & 89.98$\pm$0.56 & 71.96$\pm$1.43 & 46.25$\pm$0.92 & 55.16$\pm$1.70  & 58.95$\pm$2.87 & 51.11$\pm$3.75 & 47.31$\pm$4.48 & 69.15$\pm$3.05 & 40.57$\pm$1.44 \\ \hline
LwTR (DA)  & 74.80$\pm$1.57  & 83.01$\pm$0.41 & 75.01$\pm$3.35 & 89.79$\pm$0.28 & 70.85$\pm$1.13 & \textbf{46.78$\pm$0.54} & 54.01$\pm$2.14 & \textbf{60.86$\pm$2.89} & \textbf{53.89$\pm$3.76} & \textbf{50.20$\pm$3.77}  & \textbf{69.53$\pm$1.60} & 40.80$\pm$0.97  \\
MR (DA)    & 73.57$\pm$1.09 & 81.52$\pm$2.09 & 71.43$\pm$3.80  & 89.90$\pm$0.34  & 68.31$\pm$3.52 & 44.88$\pm$1.38 & 53.73$\pm$2.35 & 57.46$\pm$3.70  & 52.74$\pm$3.27 & 46.90$\pm$4.87  & 68.57$\pm$2.71 & 40.50$\pm$1.79  \\
SR (DA)    & 73.64$\pm$3.45 & 82.03$\pm$2.14 & 72.25$\pm$4.88 & \textbf{90.24$\pm$0.11} & 66.18$\pm$4.59 & 46.38$\pm$1.45 & 53.41$\pm$2.46 & 58.54$\pm$3.20  & 53.08$\pm$4.85 & 46.48$\pm$7.08 & 68.13$\pm$1.41 & \textbf{41.20$\pm$1.23}  \\
MELM (DA)  & 73.46$\pm$2.46 & 82.22$\pm$1.23 & 75.56$\pm$2.60  & 89.94$\pm$0.18 & 62.43$\pm$2.95 & 45.19$\pm$0.97 & 48.01$\pm$5.27  & 49.59$\pm$6.16 & 48.93$\pm$4.11 & 42.09$\pm$5.61 & 63.46$\pm$2.28 & 36.16$\pm$3.76 \\ \hline
\end{tabular}
}
\caption{$F_{1}$ scores of existing calibration methods and data augmentation methods in OntoNotes 5.0.}
\label{tab:ontonotes5_f1_scores}
\end{table*}

\begin{table}[b]
\centering
\scalebox{0.65}{
\begin{tabular}{l|cccc}
\hline
Methods & $\mathtt{EN}$ & $\mathtt{DE}$ & $\mathtt{ES}$ & $\mathtt{HI}$ \\
\hline
Baseline & 68.80$\pm$0.38 & 64.91$\pm$0.60 & 63.53$\pm$0.41 & 37.33$\pm$3.77 \\
TS & 68.51$\pm$0.52 & 64.70$\pm$0.90 & 63.41$\pm$0.45 & 37.90$\pm$2.79 \\
LS & 69.17$\pm$0.55 & 65.37$\pm$0.51 & 63.83$\pm$0.32 & 39.93$\pm$3.50 \\
MC Dropout & 68.56$\pm$0.96 & 64.70$\pm$0.87 & 63.39$\pm$0.69 & 36.38$\pm$5.89 \\ \hline
LwTR (DA) & 68.86$\pm$0.82 & 64.95$\pm$0.64 & 63.52$\pm$0.85 & 38.24$\pm$3.11 \\
MR (DA) & \textbf{69.71$\pm$0.72} & \textbf{65.37$\pm$0.57} & \textbf{64.25$\pm$0.62} & 37.53$\pm$4.03 \\
SR (DA) & 68.81$\pm$0.41 & 64.75$\pm$0.86 & 63.85$\pm$0.46 & 
\textbf{42.31$\pm$1.45} \\
MELM (DA) & 68.57$\pm$0.54 & 63.40$\pm$0.49 & 62.76$\pm$0.64 & 37.78$\pm$3.16 \\
\hline
\end{tabular}
}
\caption{$F_{1}$ scores of existing calibration methods and data augmentation methods in MultiCoNER.}
\label{tab:multiconer_EN_f1scores}
\end{table}

\section{\texorpdfstring{$F_{1}$}{F1} Scores}
\label{sec:appendix_f1_scores}
Table \ref{tab:ontonotes5_f1_scores} and \ref{tab:multiconer_EN_f1scores} show $F_{1}$ scores. Note that in many cases, data augmentation methods do not degrade predictive performance itself, but MELM often significantly degrades predictive performance in some cases, especially when the source domains are $\mathtt{nw}$ and $\mathtt{tc}$. Considering Section \ref{sec:cross-genre-calibration} and \ref{sec:cross-lingual-calibration}, MR improves calibration and uncertainty performance in many cases without degrading predictive performance.

\end{document}